\documentclass[10pt,journal,twocolumn]{IEEEtran}
\ifCLASSOPTIONcompsoc
  \usepackage[nocompress]{cite}
\else
  \usepackage{cite}
\fi

\ifCLASSINFOpdf
\else
\fi

\usepackage{times}
\usepackage{epsfig}
\usepackage{graphicx}
\usepackage{amsmath}
\usepackage{amssymb}
\usepackage{eucal,bibspacing}
\usepackage{cs}
\usepackage{mathsymb}
\usepackage{textcomp}
\usepackage{verbatim}
\usepackage{url}
\usepackage{multirow}
\usepackage[super]{nth}
\usepackage{colortbl}
\usepackage{color}
\usepackage{pbox}

\usepackage{wrapfig}
\usepackage{caption}
\usepackage[linesnumbered,algoruled,boxed,lined]{algorithm2e}

\def\bx{{\boldsymbol x}}

\let\savedalgorithm\algorithm
\let\savedendalgorithm\endalgorithm

\usepackage{algorithm}

\renewcommand{\baselinestretch}{0.96}

\begin{document}

\title{Cycle-Consistent Deep Generative Hashing for Cross-Modal Retrieval}

\author{Lin Wu, Yang Wang and Ling Shao \emph{Senior Member, IEEE}
\IEEEcompsocitemizethanks{\IEEEcompsocthanksitem Lin Wu is with The University of Queensland, St Lucia 4072, Australia (e-mail: lin.wu@uq.edu.au).

Yang Wang is with Dalian University of Technology, Dalian 116024, China (e-mail: yang.wang@dlut.edu.cn). Corresponding Author.

Ling Shao is with Inception Institute of Artificial Intelligence (IIAI), Abu Dhabi, United Arab Emirates (e-mail: ling.shao@ieee.org).\protect\\

}
}

\IEEEtitleabstractindextext{%
\begin{abstract}
In this paper, we propose a novel deep generative approach to cross-modal retrieval to learn hash functions in the absence of paired training samples through the cycle consistency loss. Our proposed approach employs adversarial training scheme to lean a couple of hash functions enabling translation between modalities while assuming the underlying semantic relationship. To induce the hash codes with semantics to the input-output pair, cycle consistency loss is further proposed upon the adversarial training to strengthen the correlations between inputs and corresponding outputs. Our approach is generative to learn hash functions such that the learned hash codes can maximally correlate each input-output correspondence, meanwhile can also regenerate the inputs so as to minimize the information loss. The learning to hash embedding is thus performed to jointly optimize the parameters of the hash functions across modalities as well as the associated generative models. Extensive experiments on a variety of large-scale cross-modal data sets demonstrate that our proposed method achieves better retrieval results than the state-of-the-arts.
\end{abstract}

}

\maketitle

\IEEEdisplaynontitleabstractindextext

\IEEEpeerreviewmaketitle

\section{Introduction}\label{sec:intro}

\begin{figure}[t]
\centering
\includegraphics[height=3cm]{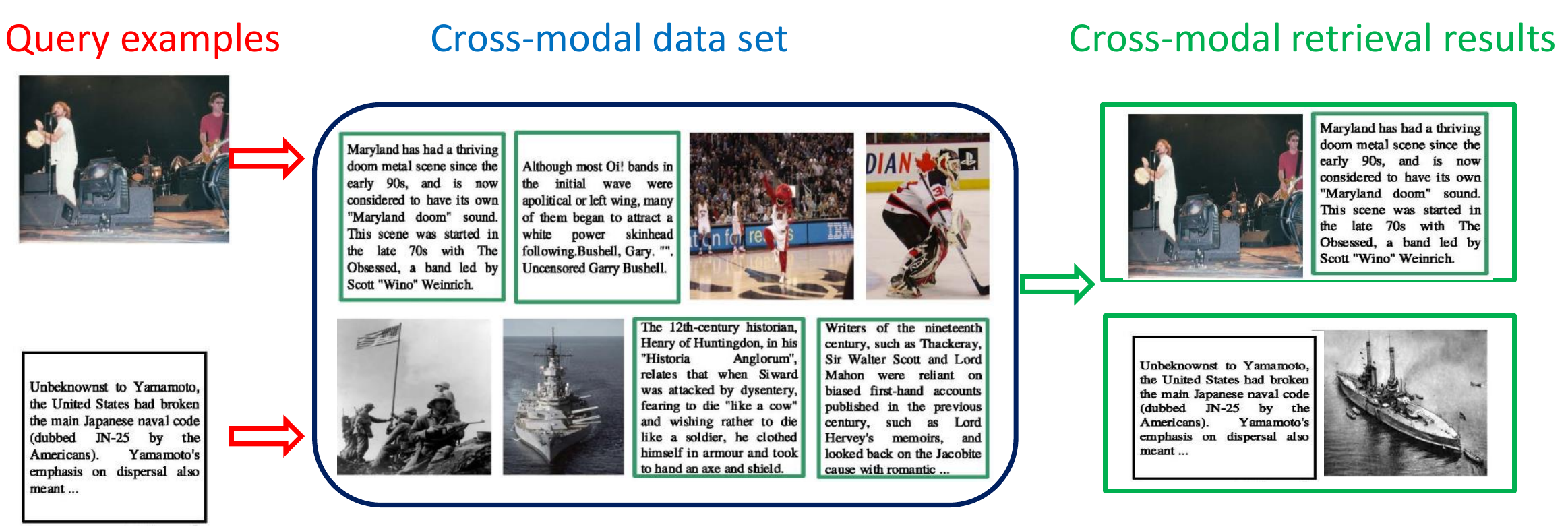}
\caption{The illustration on cross-modal retrieval with images and texts.}
\label{fig:cross-modal-example}
\end{figure}

\IEEEPARstart{T}{he} sheer volumes of big multimedia data with different modalities, including images, videos, and texts, are now mixed together and represent comprehensive knowledge to perceive the real world. It thus attracts increasing attention to approximate nearest neighbor search across different media modalities that brings both computation efficiency and search quality. Naturally, entities in correspondence from heterogenous modalities may endow semantic correlations, and it tends to entail cross-modal retrieval that returns relevant search results from one modality as response to query of another modality, e.g., retrieval of texts/images by using a query image/text, as shown in Fig.\ref{fig:cross-modal-example}.

A viable solution to large-volume cross-modal retrieval is to develop hash methods that learns compact binary codes as similar/dissimilar as possible if they have the same/different semantics. However, effective and efficient cross-modal hashing remains a big challenge due to the heterogeneity across divergent modalities \cite{Semantics-hash}, and the semantic gap between low-level features and high-level semantics. A large body of cross-modal hashing methods are proposed to learn projections from different modalities into an independent semantic embedding space with respect to characterize the model-specific relationship \cite{LBMCH,Shao-TPAMI,Wu-cyber18,Yang-TNNLS17,Yang-TIP17,Wu-PR182,Yang-TIP15,YangCIKM2013,LinMM2013,Wu-PR18,YangMM2015}. However, these shallow methods essentially learn a single pair of linear or non-linear projections to map each example into a binary code. Optimizing single projections towards each modality is suboptimal, and also the low-level descriptions on images are limited in expressing their high-level semantics. Some recent models based on deep learning are developed for cross-modal hashing \cite{DVSH,CMDVH,Multimodal-hashing,Corr-AE,CAH,DCMH}. These supervised deep models utilize semantic labels to enhance the correlation of cross-modal data wherein the feature transformations and hash functions can be jointly learned in an end-to-end manner.

\begin{figure}[t]
\centering
\includegraphics[height=3cm]{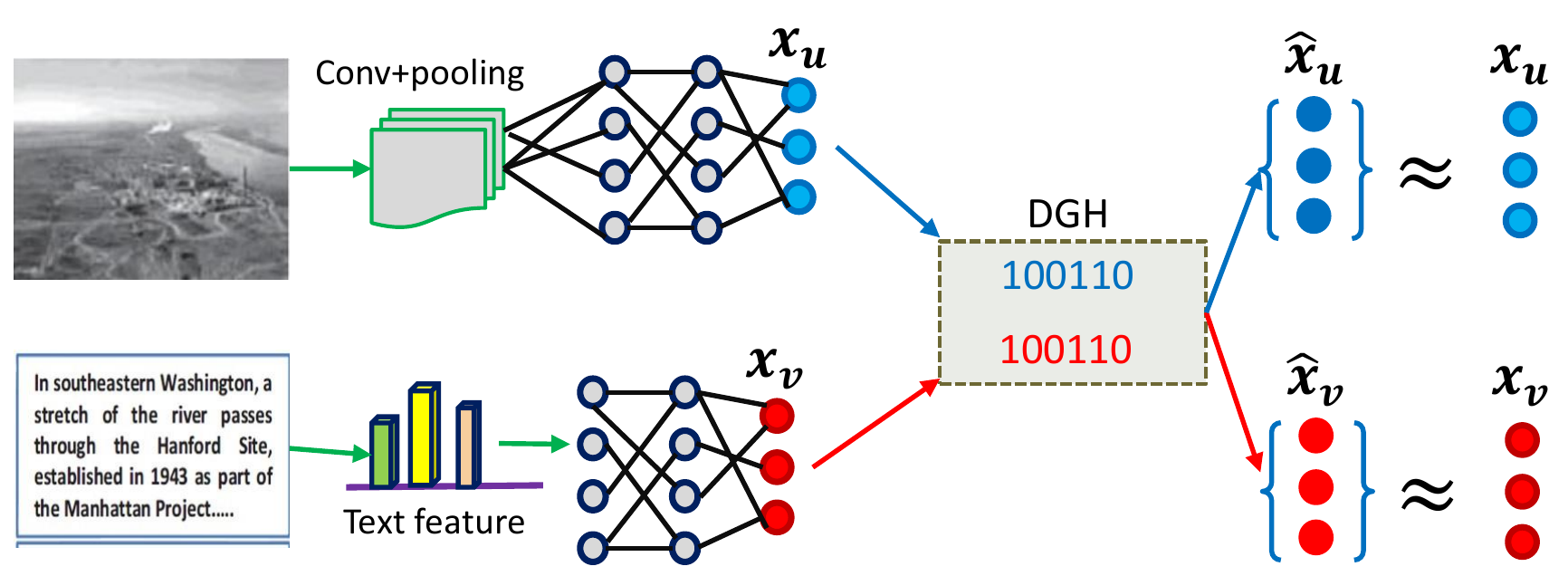}
\caption{The architecture overview of our proposed CYC-DGH for cross-modal retrieval. Given two modalities of images and texts, we learn a couple of generative hash functions in the absence of paired correspondence through the cycle-consistent adversarial learning. This is achieved by training two mappings: $\boldsymbol G: \boldsymbol X_u \rightarrow \boldsymbol X_v$ and $\boldsymbol F: \boldsymbol X_v \rightarrow \boldsymbol X_u$, which are reverse of each other, and adding a cycle consistency loss into the adversarial loss yields $\boldsymbol F(\boldsymbol G (\boldsymbol x_u)) \approx \boldsymbol x_u$, and $\boldsymbol G(\boldsymbol F (\boldsymbol x_v)) \approx \boldsymbol x_v$. The deep generative model performs both hash function learning and regeneration inputs from binary codes. See text for details.}
\label{fig:framework}
\end{figure}

It is admitted that cross-modal retrieval has been made towards substantial progression by the promotion on deep learning models. We remark that there are two major challenges remained open to be addressed: First, the cross-modal hashing is performed to learn the mapping between an input image/text and an output text/image using a training set of labeled \textit{aligned} pairs. The supervision of paired correspondence is to enhance the correlation of cross-modal data such that the hashing can be guided by preserving the semantics. For instance, Zhang et al. \cite{SCM} performed semantic correlation maximization using label information to learn modality-specific transformations. However, for many realistic cases, paired training data will not be available. Even labeling is feasible, deep hashing models trained on limited amount of labeled samples are inclined to be over-fitting, and thus the generalization is not guaranteed. Second, transforming an input which typically in high-dimension into its binary codes will inevitably cause information loss. Existing hashing methods unidirectionally learn hash functions from inputs to hash codes. However, the hash codes can also be used to regenerate the inputs \cite{SGH,Reconstruction-hashing}, which should be exploited to characterize the goodness of hash codes and regenerating inputs through hash codes provides a principle to minimize the information loss during the hash embedding.

Recent works have shown that generative adversarial networks combined with cycle-consistency constraints are surprisingly effective at mapping images between domains, even without the use of aligned image pairs \cite{CycleGAN,CyCADA}. We therefore seek an algorithm that can learn to translate the binary codes between modalities without paired image-text examples. As shown in \cite{CycleGAN}, there is some underlying relationship between the two modalities, for example, they are two different renderings of the same underlying scene and the relationship can be learnt by exploiting the supervision at the level of the sets: an adversary objective can be trained to induce an output distribution over the target modality that matches its empirical distribution. The optimal mapping thereby translates the source modality to a new modality distributed identically to the new one, whereas a cycle-consistent constraint is needed to guarantee an individual input and output are paired up in an meaningful way.

\subsection{Our Approach: CYC-DGH}

For addressing the two key issues in large-scale cross-modal retrieval, we propose Cycle-Consistent Deep Generative Hashing (CYC-DGH), which aims to produce hash embedding in the absence of paired training correspondence. The basic idea of the proposed approach is shown in Fig.\ref{fig:framework}.

Specifically, our goal is to learn a couple of hash mappings that can translate between domains \footnote{In this paper, we use the terms of domain and modality alternatively without discrimination.} without paired input-output examples. We assume there is some underlying relationship between the domains, for example, they are images and texts of the same semantic meaning, and seek that relationship. Although we lack supervision in the form of paired examples, we can exploit supervision at the level of sets: given one set of images in modality $\boldsymbol X_u$ and a different set of texts in modality $\boldsymbol X_v$. We may train a mapping $\boldsymbol G: \boldsymbol X_u \rightarrow \boldsymbol X_v$ such that the output $\hat{\boldsymbol x}_v =\boldsymbol G(\boldsymbol x_u)$, $\boldsymbol x_u \in \boldsymbol X_u$, is indistinguishable from texts $\boldsymbol x_v \in \boldsymbol X_v$ by an adversary trained to classify $\hat{\boldsymbol x}_v$ apart from $\boldsymbol x_v$. The optimal $\boldsymbol G$ thereby translates the modality $\boldsymbol X_u$ into a modality $\hat{\boldsymbol X}_v$ distributed identically to $\boldsymbol X_v$. As a result, the modality gap can be reduced effectively. However, such a translation is highly under-constrained, and does not guarantee that an individual input $\boldsymbol x_u$ and an output $\boldsymbol x_v$ are matched in a meaningful way. In fact, there could be infinitely mappings $\boldsymbol G$ that will induce the same distribution over $\hat{\boldsymbol x}_v$. Towards this end, we exploit the property that the translation should be ``cycle-consistent" \cite{CycleGAN}, in the sense that if we translate, e.g., a sentence from English to French, and then translate it back from French to English, we should arrive back at the original sentence. In the case of cross-modality, if we have a domain translator $\boldsymbol G: \boldsymbol X_u \rightarrow \boldsymbol X_v$, and another translator $\boldsymbol F: \boldsymbol X_v \rightarrow \boldsymbol X_u$, then both $\boldsymbol G$ and $\boldsymbol F$ should be reverse of each other, and thus bijections. Hence, we train the mappings $\boldsymbol G$ and $\boldsymbol F$ simultaneously by combining a cycle consistency loss \cite{3D-cycle} with adversarial losses on modalities $\boldsymbol X_u$ and $\boldsymbol X_v$ that encourages $\boldsymbol F(\boldsymbol G (\boldsymbol x_u)) \approx \boldsymbol x_u$, and $\boldsymbol G(\boldsymbol F (\boldsymbol x_v)) \approx \boldsymbol x_v$.

To allow the regeneration from hash codes to the inputs so as to minimize the information loss, we decompose the mappings $\boldsymbol G$ into: $\boldsymbol G: \boldsymbol x_u \rightarrow \boldsymbol H_u \rightarrow \boldsymbol P_u \rightarrow \boldsymbol x_v$, where $\boldsymbol H$ denotes the binary code learning and $\boldsymbol P$ is the reverse process of regenerating inputs from binary codes. Finally, the cycle consistent training can bring each input $\boldsymbol x_u$ back to itself through the generative model: $\boldsymbol x_u \rightarrow \boldsymbol G(\boldsymbol x_u) \rightarrow \boldsymbol F( \boldsymbol G(\boldsymbol x_u)) \approx \boldsymbol x_u$. The similar translation can also be applied on $\boldsymbol F$, which is omitted for simplicity. The proposed generative model which captures both the encoding of binary codes from the input and the decoding of input from binary codes, provides a principled hash learning framework, where the information loss during hash embedding is minimized. Therefore, the generated codes can compress the input data with maximum preserving of its information on its own domain as well as the relationship of samples from different modalities. And also the modality gap between the hash functions are reduced. Prior works on binary auto-encoders \cite{Corr-AE} and deep generative models \cite{TUCH,CMDVH} also takes a generative view of hashing but still requires the correlation from paired samples. Generative hashing is introduced in \cite{SGH} where the hash functions can be learned through minimum description length. However, their algorithm is limited in single-modality setting.

\subsection{Contributions}
The main contributions can be summarized as follows.
\begin{itemize}
  \item Cross-modal adversarial mechanism is presented to perform a novel adversarial training over the cross-modal scenario, which can deal with the heterogeneity gap between different modalities by effectively modeling the data distribution.
  \item Cycle-consistent is introduced into the cross-modal adversarial training to enable the learning of hash functions in the absence of paired training samples.
  \item Deep generative models are proposed to learn to regenerate the input from binary codes, which is coupled with hash function learning, and thus demonstrated to be able to reconstruct the inputs so as to minimize the information loss in hash embedding.
\end{itemize}

The rest of this paper is organized as follows: We briefly introduce the related works on generative adversarial networks, cross-modal hashing methods as well as image-to-image translation literature in Section \ref{sec:related}. Section \ref{sec:approach} presents our proposed CYC-DGH approach. Section \ref{sec:exp} includes the experiments of cross-modal retrieval conducted on three cross-modal data sets with result analyses. Finally Section \ref{sec:con} concludes the paper.

\section{Related Work}\label{sec:related}

\subsection{Generative Adversarial Networks}
Recently, generative adversarial networks (GANs) have been proposed to estimate a generative model by an adversarial training process \cite{GAN}, and GANs-based networks can be used to generate new data such as image synthesis \cite{DCGANs} and video prediction \cite{Interact-Video}. In \cite{DSH-GANs}, semi-supervised GANs are exploited to deal with few labeled training data by producing synthetic examples conditioning on class labels. Due to the strong ability of GAN in modeling data distribution, GANs have been utilized to model the joint distribution over the heterogenous data of different modalities \cite{CM-GANs}. However, Peng et al. \cite{CM-GANs} aim to use GANs to learn the common representation and boost the cross-modal correlation learning, which is a completely different goal of our method to learning hash functions without the dependence of paired training samples across modalities. Some advanced models are developed to transfer cross-view sentence-image retrieval problem into a single-view problem. For instance, an end-to-end differentiable architecture from the character level to pixel level is proposed using model conditioned on text descriptions to achieve sentence text to corresponding image synthesis \cite{Generate-txt-img}. Inspired by this idea, Zhao et al. \cite{TUCH} turn cross-view hashing into single-view by generating fake images from text feature and jointly learn hash functions.

\subsection{Deep Cross-Modal Hashing}

Cross-modal multimedia retrieval performs the task of retrieving text documents by using a given query image and vice versa. The objective for many cross-modal methods is learn a common subspace between images and texts to model the correlations \cite{Cross-modal-MM2010,Coupled-modal,Joint-feature-subspace}. For example, in the literature canonical component analysis (CCA) is used to map both text documents and images into a latent space \cite{Cross-modal-MM2010}. Wang \textit{et al.} \cite{Coupled-modal} learn a coupled feature space to select the most relevant and discriminative features for cross-modal matching.

In order to suit large-scale search, cross-modal hashing becomes a more desirable choice for efficiency \cite{LBMCH}. The majority of cross-modal hash methods can be classified into two types: unsupervised (CVH \cite{CVH}, LSSH \cite{LSSH}) and supervised (SCM \cite{SCM}, SePH \cite{Semantics-hash}). Unsupervised methods utilize co-occurrence information such that only the image-text pairs which occur in the same article are considered to be of similar semantic. For example, Zhou et al. \cite{LSSH} obtain a unified binary from a latent space learning method by using sparse coding and matrix factorization in the common space. On the other hand, supervised methods utilize semantic labels to enhance the correlation of cross-modal pairs. For instance, a semantic correlation maximization (SCM) is performed to use label information to learn a modality-specific transformation which can maximize the correlation between modalities. However, these studies are in shallow form in the sense that they only perform a single-layer of linear or non-linear transformation.

While there are a large body of methods that perform deep learning for cross-modal retrieval by integrating feature learning and hashing coding into end-to-end trainable frameworks \cite{Deep-binary,Deep-sketch-hashing,DVSH,CMDVH}. For instance, Cao et al. \cite{DVSH} learned a visual semantic fusion network with cosine hinge loss to obtain the binary codes and learned modality-specific networks to obtain the hash functions. To make the hashing network suitable for out-of-sample extension, a cross-modal deep variational hashing method (CMDVH) \cite{CMDVH} is proposed to reformulate the modality-specific hashing networks into a generative form. They introduce a set of latent variables that are modeled similar to the inferred unified binary codes for each paired image/text sample through a plausible log likelihood criterion. However, this work still performs binary code inference from labeled training pairs to seek a common hamming space. In contrast, our approach is effective in reducing the modality gap in the absence of pairs by the merit of cycle-consistency loss in adversarial training to encourage the modality alignment. HashGAN \cite{HashGAN}proposed an adversarial hashing network with attention mechanism to enhance the measurement of content similarities by selectively focusing on informative parts of multi-modal data. Shen et al. \cite{ZSIH} presented a zero-shot sketch-image hashing (ZSIH) model to address the never-seen observation in training. The modality gap and semantic correlation between sketch-image can be enhanced by using a Kronecker fusion layer and graph convolution. They also formulate a generative hashing scheme in reconstructing semantic knowledge representations for zero-shot retrieval. However, the ZSIH model is dependent on paired sketch-image for training which is somewhat limited in practical situations.

\begin{figure*}[!t]
\centering
\includegraphics[height=4cm]{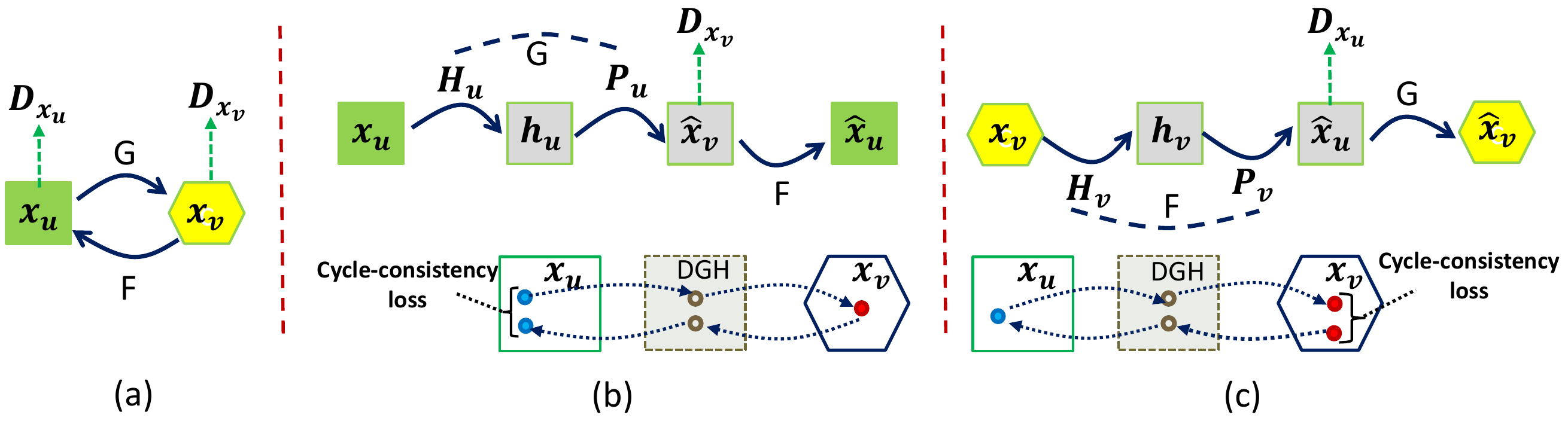}
\caption{The proposed cycle-consistent deep generative hashing (CYC-DGH) for cross-modal retrieval. (a) The model of CYC-DGH couples two mappings: $\boldsymbol G: \boldsymbol x_u \rightarrow \boldsymbol x_v$ and $\boldsymbol F: \boldsymbol x_v \rightarrow \boldsymbol x_u$ as well as associated adversarial discriminators $\boldsymbol D_{\boldsymbol x_v}$ and $\boldsymbol D_{\boldsymbol x_u}$. The two mappings are decomposed into the binary code generation and the reverse process of regenerating inputs from binary codes: $\boldsymbol G: \boldsymbol x_u \rightarrow \boldsymbol H_u \rightarrow \boldsymbol P_u \rightarrow \boldsymbol x_v$ and $\boldsymbol F: \boldsymbol x_v \rightarrow \boldsymbol H_v \rightarrow \boldsymbol P_v \rightarrow \boldsymbol x_u$. To regulate the mappings, two cycle-consistent losses are introduced: (b) forward $\boldsymbol x_u \rightarrow G(\boldsymbol x_u) \rightarrow \boldsymbol F( \boldsymbol G(\boldsymbol x_u)) \approx \hat{\boldsymbol x}_u$, and (c) backward $\boldsymbol x_v \rightarrow F(\boldsymbol x_v) \rightarrow \boldsymbol G(\boldsymbol F(\boldsymbol x_v)) \approx \hat{\boldsymbol x}_v$. }
\label{fig:cycle-loss}
\end{figure*}

\subsection{Image-to-Image Translation}

The task of image-to-image translation is to translate one possible representation of a
scene into another, given sufficient training data. More recent approaches use a collection of input-output examples to learn parametric translation function based on CNNs \cite{Fully-CNN}. For instance, the ``pix2pix" framework \cite{Pix2Pix} investigates a conditional generative adversarial network to learn a mapping from input to output images. The idea of conditional adversarial nets is a general-purpose solution which has been applied into various tasks such as generating photographs from sketches \cite{Scribbler}. However, these works learn the mapping with dependency on paired training examples. To tackle with unpaired setting, some methods are proposed to relate two data domains. For instance, CoGAN \cite{Coupled-GAN} and cross-modal scene networks \cite{Cross-scene-network} use a weight-sharing strategy to learn a common representation across domains. Recent studies show that higher-order cycle consistency can be used in depth estimation \cite{Depth-left-right} and co-segmentation \cite{co-segmentation-consistent} in which a cycle consistency loss can be used as a way of using transitivity to supervise the CNN training. And Zhu et al. \cite{CycleGAN} introduce a similar loss to push the mappings to be consistent with each other. Concurrent with \cite{CycleGAN}, Yi et al. \cite{Dual-GAN} use a similar objective for unpaired image-to-image translation, inspired by the dual learning in machine translation. Our work is also inspired by the cycle-consistency loss, which is introduced into the cross-modal hashing learning to alleviate the pairing on training samples. In addition, we extend the cycle-consistency pipeline by enforcing the hash mappings to be translated without requiring paired training samples.
\section{Cycle-Consistent Deep Generative Hashing}\label{sec:approach}


In this section, we propose an end-to-end deep architecture for cross-modal hashing such that we are able to maximize the correlation between the two modalities in the absence of paired training examples. The network composes of generative hash functions in regards to two modalities for generating the binary codes from input as well as reversely generating input from binary codes, and mapping functions between two modalities without paired correspondence.

Let $\boldsymbol X_u= [\boldsymbol x_{u1}, \boldsymbol x_{u2},\cdots, \boldsymbol x_{uN}]\in \mathbb{R}^{d_u \times N}$ and $\boldsymbol X_v= [\boldsymbol x_{v1}, \boldsymbol x_{v2},\cdots, \boldsymbol x_{vN}]\in \mathbb{R}^{d_v \times M}$ be the training samples from different modalities, where $u$ and $v$ denote two different modalities, and each sample $\boldsymbol x_u$ ($\boldsymbol x_v$) is produced by the neural network with parameters $f_u (\cdot)$ ($f_v(\cdot)$). Our objective contains three types of terms: adversarial losses \cite{GAN} for matching the distribution of generated samples to the data distribution in the target modality; the cycle-consistency losses \cite{CycleGAN} to present the learned mappings $\boldsymbol G$ and $\boldsymbol F$ from contradicting each other; the reconstruction loss of reconstructing the input from the binary codes. In the following, we present the respective losses and show how to perform optimizations in the proceeding subsections.

\subsection{Adversarial Loss}

We denote the data distribution as $\boldsymbol x_u \sim p_{data} (\boldsymbol x_u)$ and $\boldsymbol x_v \sim p_{data} (\boldsymbol x_v)$. As shown in Fig. \ref{fig:cycle-loss} (a), for the cross-modal correlation, our model includes two mappings $\boldsymbol G: \boldsymbol x_u \rightarrow \boldsymbol x_v$ and $\boldsymbol F: \boldsymbol x_v \rightarrow \boldsymbol x_u$. In addition, we introduce two adversarial discriminators: $\boldsymbol D_{\boldsymbol x_u}$ and $\boldsymbol D_{\boldsymbol x_v}$, where $\boldsymbol D_{\boldsymbol x_u}$ aims to distinguish between images $\{\boldsymbol x_u\}$ and translated texts $\{\boldsymbol F(\boldsymbol x_v)\}$; in the same way, $\boldsymbol D_{\boldsymbol x_v}$ aims to discriminate between $\{ \boldsymbol x_v\}$ and $\{\boldsymbol G(\boldsymbol x_u)\}$. For the mapping function $\boldsymbol G: \boldsymbol x_u \rightarrow \boldsymbol x_v$ and its discriminator $\boldsymbol D_{\boldsymbol x_v}$, the objective can be expressed as:
\begin{equation}\label{eq:adversarial}
\begin{split}
     & \mathcal{L}_{GAN} (\boldsymbol G, \boldsymbol D_{\boldsymbol x_v}, \boldsymbol x_u, \boldsymbol x_v)= \mathbb{E}_{\boldsymbol x_v \sim p_{data} (\boldsymbol x_v)} [\log \boldsymbol D_{\boldsymbol x_v}(\boldsymbol x_v)] \\
     & + \mathbb{E}_{\boldsymbol x_u \sim p_{data} (\boldsymbol x_u)} [\log ( 1 - \boldsymbol D_{\boldsymbol x_v}( \boldsymbol G (\boldsymbol x_u) ))],
\end{split}
\end{equation}
where $\boldsymbol G$ attempts to generate images $\boldsymbol G (\boldsymbol x_u)$ that look similar to items from domain $\boldsymbol X_v$, while $\boldsymbol D_{\boldsymbol x_v}$ aims to distinguish between translated samples $\boldsymbol G (\boldsymbol x_u)$ and real samples $\boldsymbol x_v$. $\boldsymbol G$ aims to minimize the objective against an adversary $\boldsymbol D_{\boldsymbol x_v}$ that tries to maximize it, i.e., $\min_{\boldsymbol G} \max_{\boldsymbol D_{\boldsymbol x_v}} \mathcal{L}_{GAN} (\boldsymbol G, \boldsymbol D_{\boldsymbol x_v}, \boldsymbol x_u, \boldsymbol x_v)$. Similarly, an adversarial loss is introduced for the mapping function $\boldsymbol F: \boldsymbol x_v \rightarrow \boldsymbol x_u$ and its discriminator $\boldsymbol D_{\boldsymbol x_u}$: i.e., $\min_{\boldsymbol F} \max_{\boldsymbol D_{\boldsymbol x_u}} \mathcal{L}_{GAN} (\boldsymbol F, \boldsymbol D_{\boldsymbol x_u}, \boldsymbol x_v, \boldsymbol x_u)$.

\subsection{Cycle-Consistency Loss}

One characteristic of cross-modal retrieval is to minimize the modality discrepancy, and thus the semantics can be consented and represented by objects in different modalities. Existing methods typically deploy the paired correspondence to supervise the learning of a common subspace between images and text \cite{Coupled-modal,LBMCH} or the translation between the two domains \cite{TUCH}. However, manually labelling pairs is not viable in web-scale multimedia retrieval. In addition, learning a mapping between a number of specific image-text pairs is limited in producing a general-purpose solution of capturing high-level correspondences in many vision tasks. To this end, cycle-consistency loss \cite{CycleGAN} is introduced to learn mappings translated across two domains without paring constraint. This consistent loss is motivated to regulate the adversarial training to guarantee the learned function can map an individual input $\boldsymbol x_{ui}$ to a desired output $\boldsymbol x_{vi}$.

In our case, the two mappings $\boldsymbol G$ and $\boldsymbol F$ are composed of binary code generation ($\boldsymbol H_{\ast}: \boldsymbol x_{\ast} \rightarrow \boldsymbol h_{\ast} \in \{0,1\}^{K}$, where $\ast=\{u, v\}$, $\hat{\ast}\neq \ast$) and the reverse process of generating inputs from binary codes ($\boldsymbol P_{\ast}: \boldsymbol h_{\ast} \rightarrow \boldsymbol x_{\hat{\ast}}$, where $\ast=\{u, v\}$), that are, $\boldsymbol G: \boldsymbol x_u \rightarrow \boldsymbol H_u \rightarrow \boldsymbol P_u \rightarrow \boldsymbol x_v$ and $\boldsymbol F: \boldsymbol x_v \rightarrow \boldsymbol H_v \rightarrow \boldsymbol P_v \rightarrow \boldsymbol x_u$, as shown in Fig.\ref{fig:cycle-loss} (b) and (c), respectively. Thus, the learned mapping functions should be cycle-consistent: as shown in Fig. \ref{fig:cycle-loss} (b), for each input $\boldsymbol x_u$ from modality $\boldsymbol X_u$, the input translation cycle should be able to bring $\boldsymbol x_u$ back to the original input through the generative hamming space ($\{\boldsymbol H_{\ast}, \boldsymbol P_{\ast} \}$), i.e., $\boldsymbol x_u \rightarrow G(\boldsymbol x_u) \rightarrow \boldsymbol F( \boldsymbol G(\boldsymbol x_u)) \approx \hat{\boldsymbol x}_u$. Similarly, as illustrated in Fig. \ref{fig:cycle-loss} (c), for each input $\boldsymbol x_v$ from modality $\boldsymbol X_v$, $\boldsymbol G$ and $\boldsymbol F$ should also satisfy: $\boldsymbol x_v \rightarrow F(\boldsymbol x_v) \rightarrow \boldsymbol G(\boldsymbol F(\boldsymbol x_v)) \approx \hat{\boldsymbol x}_v$.

\begin{equation}\label{eq:cycle-loss}
\begin{split}
     &   \mathcal{L}_{cyc}(\boldsymbol G,\boldsymbol F)=\mathbb{E}_{\boldsymbol x_u \sim p_{data} (\boldsymbol x_u)} \left[ || \boldsymbol F(\boldsymbol G(\boldsymbol x_u) - \boldsymbol x_u)||_1\right] \\
     &  + \mathbb{E}_{\boldsymbol x_v \sim p_{data} (\boldsymbol x_v)} \left[ || \boldsymbol G(\boldsymbol F(\boldsymbol x_v) - \boldsymbol x_v)||_1\right].
\end{split}
\end{equation}

\subsection{Deep Generative Hashing}

\subsubsection{Generative Model}
In our case, we introduce a generative model that defines the likelihood of generating input $\boldsymbol x_v$ given the binary code in its correspondence $\boldsymbol h_u$, that is, $\boldsymbol P_u: \boldsymbol h_u \rightarrow \boldsymbol x_v$, denoted as $p(\boldsymbol x_v| \boldsymbol h_u)$; and we also have $\boldsymbol P_v: \boldsymbol h_v \rightarrow \boldsymbol x_u$, denoted as $p(\boldsymbol x_u| \boldsymbol h_v)$. This cross-modal generative hashing is to ensure objects in different modalities are translated through their hashing codes, and thus semantic consistence is achieved even without the paired constraint. $\boldsymbol P_\ast$ are also referred as decoding functions.

Inspired by the recent stochastic generative hashing \cite{SGH}, we use the simple Gaussian distribution to model the generation of $\boldsymbol x$ given $\boldsymbol h$ (for simplicity, we omit the subscripts), which is defined as:
\begin{equation}\label{eq:gaussian}
  p(\boldsymbol x, \boldsymbol h) = p(\boldsymbol x | \boldsymbol h) p(\boldsymbol h), p(\boldsymbol x | \boldsymbol h)= \mathcal{N}(\boldsymbol U \boldsymbol h, \rho^2 \boldsymbol I),
\end{equation}
where $\boldsymbol U =\{\boldsymbol u_i\}_{i=1}^K$, $\boldsymbol u_i\in \mathbb{R}^d$ is a code book with $K$ code words. The prior $p(\boldsymbol h) \sim \mathcal{B}(\theta)=\Pi_{i=1}^K \theta_i^{\boldsymbol h_i} (1-\theta_i)^{1-\boldsymbol h_i}$ is modeled as the multivariate Bernoulli distribution on the hash codes, where $\theta=[\theta_i]_{i=1}^K \in [0,1]^K$. This intuition can be interpreted as an additive model which reconstructs the input $\boldsymbol x$ by summing the selected columns of $\boldsymbol U$ given $\boldsymbol h$, with a Bernoulli prior on the distribution of hash codes. The joint distribution can be formulated as:
\begin{equation}\label{eq:joint}
  p(\boldsymbol x, \boldsymbol h) \propto \exp \left( \frac{1}{2\rho^2} || \boldsymbol x-\boldsymbol U^T \boldsymbol h ||_2^2- (\log \frac{\theta}{1-\theta})^T \boldsymbol h\right).
\end{equation}
It has been shown that the Gaussian reconstruction error $|| \boldsymbol x-\boldsymbol U^T \boldsymbol h ||_2^2$ is a surrogate for Euclidean neighborhood preservation \cite{SGH}, and thus this generative model is able to preserve the local neighborhood structure of the input $\boldsymbol x$ when the Frobenius norm of $\boldsymbol U$ is bounded, that is, minimizing the Gaussian reconstruction error $-\log p(\boldsymbol x| \boldsymbol h)$ will leads to the Euclidean neighborhood preservation. This property is critical to cross-modal retrieval in the sense that the modality-specific local neighborhood structure of data objects can be well characterized.

\subsubsection{Encoding Model ($\boldsymbol H_{\ast}: \boldsymbol x_{\ast} \rightarrow \boldsymbol h_{\ast}$)}

Directly computing the posterior $p(\boldsymbol h| \boldsymbol x)=\frac{p(\boldsymbol x, \boldsymbol h)}{p(\boldsymbol x)}$, and seeking the maximum a posterior (MAP) solution to the posterior involves solving an expensive integer programming subproblem. Some recent studies on variational auto-encoder \cite{AEV,Mnih2014} show that this difficulty can be avoided by fitting an approximate inference model, $q(\boldsymbol h | \boldsymbol x)$, to approximate the exact posterior of the encoding function $p(\boldsymbol h| \boldsymbol x)$. Thus, the encoding function can be re-parameterized as
\begin{equation}\label{eq:encode-function}
  q(\boldsymbol h | \boldsymbol x) = \Pi_{k=1}^K q(\boldsymbol h_k=1| \boldsymbol x)^{\boldsymbol h_k} q(\boldsymbol h_k=0 | \boldsymbol x)^{1-\boldsymbol h_k},
\end{equation}
where $\boldsymbol h = [\boldsymbol h_k]_{k=1}^K \sim \mathcal{B}(\sigma(\boldsymbol W^T \boldsymbol x))$ is the linear parametrization where $\boldsymbol W=[\boldsymbol w_k]_{k=1}^K$. Hence, we can have
\begin{equation}\label{eq:hash-function}
  p(\boldsymbol h | \boldsymbol x)=\arg\max_{\boldsymbol h} q(\boldsymbol h | \boldsymbol x)=\frac{sign(\boldsymbol W^T \boldsymbol x) + 1}{2}.
\end{equation}

\subsection{Training Objective}

Our full training objective is formulated to be:

\begin{equation}\label{eq:objective}
\begin{split}
     & \mathcal{L} (\boldsymbol G, \boldsymbol F, \boldsymbol D_{\boldsymbol x_v}, \boldsymbol D_{\boldsymbol x_u} , \boldsymbol H_{\ast})= \mathcal{L}_{GAN} (\boldsymbol G, \boldsymbol D_{\boldsymbol x_v}, \boldsymbol x_u, \boldsymbol x_v) \\
     & + \mathcal{L}_{GAN} (\boldsymbol F, \boldsymbol D_{\boldsymbol x_u}, \boldsymbol x_v, \boldsymbol x_u) + \lambda \mathcal{L}_{cyc}(\boldsymbol G, \boldsymbol F) \\
     & + D_{KL} (q(\boldsymbol h_{\ast}| \boldsymbol x_{\ast}) || p( \boldsymbol h_{\ast} | \boldsymbol x_{\ast})) + L(\Theta_{\ast}; \boldsymbol x_{\ast})
\end{split}
\end{equation}
where $\ast=\{u,v\}$ denotes the modality, $L(\Theta_{\ast}; \boldsymbol x_{\ast})=\mathbb{E}_{q(\boldsymbol h_{\ast}| \boldsymbol x_{\ast})}\left[ -\log q(\boldsymbol h_{\ast}| \boldsymbol x_{\ast}) + \log p(\boldsymbol x_{\ast},\boldsymbol h_{\ast} ) \right]$, $\lambda$ controls the relative importance of the two sub-objectives in cycle-consistence training, and $\Theta_{\ast}=\{ \boldsymbol W_{\ast}, \boldsymbol U_{\ast}, \rho_{\ast}, \beta_{\ast}:=\log \frac{\theta}{1-\theta}\}$. $\boldsymbol U_{\ast}, \rho_{\ast}, \beta_{\ast}:=\log \frac{\theta}{1-\theta}$ are the parameters of the generative model $p(\boldsymbol x_{\ast},\boldsymbol h_{\ast})$ and $\boldsymbol W_{\ast}$ comes from the encoding function Eq. \eqref{eq:encode-function}. Our model can be intuitively viewed as training two auto-encoders in which we learn one auto-encoder in the form of $\boldsymbol F \circ \boldsymbol G: \boldsymbol X_u \rightarrow \boldsymbol X_u$, jointly with another $\boldsymbol G \circ \boldsymbol F: \boldsymbol X_v \rightarrow \boldsymbol X_v$. One difference is our auto-encoder training has an internal structure, that is, it maps an image/text to itself via an intermediate representation that is a translation of the image/text to another domain. This is similar to the adversarial auto-encoders \cite{AAE} where an adversarial loss is used to train an auto-encoder to match an arbitrary target distribution, whereas in our case the target distribution for the auto-encoder $\boldsymbol X_u \rightarrow \boldsymbol X_u$ is the modality of $\boldsymbol X_v$.

However, further closer examination on the training objective suggests that it is able to directly compute the gradients w.r.t parameters of $p(\boldsymbol x | \boldsymbol h)$. In particular, it is difficult to compute the stochastic gradients w.r.t $\boldsymbol W$ because it depends on the stochastic binary variables $\boldsymbol h$. In order to back-propagate the discrete stochastic variables, we follow the SGH \cite{SGH} to adopt an approximation to the gradient w.r.t $\boldsymbol W$, which is derived based on distributional derivatives. Specifically, a stochastic neuron is introduced for re-parameterizing the Bernoulli variable $\boldsymbol h_k(z)$ with $z\in (0,1)$, which is defined as

\begin{equation}\label{eq:stochastic-neuron}
\hat{\boldsymbol h}(z,\xi) = \begin{cases} 1  & \mbox{if } z \geq \xi \\
0 & \mbox{if } z < \xi \end{cases}
\end{equation}
where $\xi \sim \mathcal{U}(0,1)$ is random variable. Hence, the stochastic neuron can be used to re-parameterize the binary variables $\boldsymbol h$ by replacing $[\boldsymbol h_k]_{k=1}^K (\boldsymbol x) \sim \mathcal{B}(\sigma (\boldsymbol w_k^T \boldsymbol x))$ with $[\hat{\boldsymbol h_k} (\sigma (\boldsymbol w_k^T \boldsymbol x), \xi_k)]_{k=1}^K$. Due to the discontinuity of the stochastic neuron $\hat{\boldsymbol h} (z,\xi)$, a more generalized distributional derivative \cite{Grubb-2008} can be computed instead of computing the standard Stochastic Gradient Descent.

\subsection{Training Details}

Our deep architecture and experiments are implemented on the open-source Torch7 framework. We take the text encoding model \cite{Generate-txt-img} to extract text features. The fully-connected layers on top of text features are set to be $[1000\rightarrow 500 \rightarrow 200]$, $[11500\rightarrow 500 \rightarrow 200]$, and $[10 \rightarrow 100\rightarrow 200]$ for the COCO, IAPR TC-12, and Wiki, respectively. For the generator on images, we adopt the architecture from \cite{Style-transfer} which has shown promising results for style transfer. This generative network contains two stride-2 convolutions, several residual blocks, and two fractionally-stride convolutions with stride of $\frac{1}{2}$. We use 6 blocks for all training images. For the discriminator architecture, following \cite{CycleGAN} we use $70\times 70$ PatchGAN \cite{PatchGAN}, which aims to classify whether $70\times 70$ overlapping image patches are real or fake. The patch-level discriminator has fewer parameters than a full-image discriminator, and it can be applied to arbitrarily-sized images.

As suggested by \cite{CycleGAN}, two strategies can employed to stabilize the network training. First, for the loss of $\mathcal{L}_{GAN}$, the negative log likelihood objective is replaced by a least-square loss. In specific, for a GAN loss $\mathcal{L}_{GAN} (\boldsymbol G, \boldsymbol D_{\boldsymbol x_v}, \boldsymbol x_u, \boldsymbol x_v)$, we train the $\boldsymbol G$ to minimize $\mathbb{E}_{\boldsymbol x_v \sim p_{data} (\boldsymbol x_v)} [ (\boldsymbol D_{\boldsymbol x_v}(\boldsymbol x_v)-1)^2]$, and train the $\boldsymbol D$ to minimize $\mathbb{E}_{\boldsymbol x_u \sim p_{data} (\boldsymbol x_u)} [ ( \boldsymbol D_{\boldsymbol x_v}(\boldsymbol x_u) -1)^2] + \mathbb{E}_{\boldsymbol x_v \sim p_{data} (\boldsymbol x_v)} [ (\boldsymbol D_{\boldsymbol x_v}(\boldsymbol x_v))]$. Second, to reduce the model oscillation, the discriminator is updated using a history of generated images/texts rather than the ones produced by the latest generative networks. In all experiments, we set $\lambda=10$ in Eq.\eqref{eq:objective}, and Adam solver \cite{Adam-solver} is uses with a batch size of 1. The networks are trained with a learning rate of 0.0002, which is maintained for the first 100 epoches, and linearly decayed the rate to zero over the next 100 epoches.

\section{Experiments}\label{sec:exp}

In this section, we conduct extensive experiments to evaluate the efficiency of the proposed CYC-DGH against state-of-the-arts on three widely-used benchmark datasets.

\subsection{Data Preparation}

\begin{itemize}
\item \textbf{Microsoft COCO} \cite{Microsoft-COCO} The recent release of the data set contains 82,783 training images and 40,137 validation images. For each image, the data set provides at least five sentences annotations, belonging to 80 most frequent categories as ground truth labels. Some images that have no category information are removed from training set, and thus we get 82,081 training images.
\item \textbf{IAPR TC-12} \cite{IAPR-TC} This data set consists of 20,000 images collected from a wide variety of domains, such as sports and actions, people, animals, cities, landscapes, and so on. Each image has at least 1 sentence annotations as well as category annotations generated from segmentation with 275 concepts. Following the setting of TUCH \cite{TUCH}, we select images with 22 most frequent concept tags, and thus a new training set with 18,673 images are formed.
\item \textbf{Wiki} \footnote{http://www.svcl.ucsd.edu/projects/crossmodal/} This data set contains 2,866 Wikipedia documents, where each document contains a single image and a corresponding text of at least 70 words. These documents are categorized into 10 semantic classes, and each document is from one class. Each text is represented by a 10-dimensional feature vector computed from the Latent Dirichelet Allocation model. We randomly select 75\% documents from this data set as database and the rest a query samples.
\end{itemize}

\subsection{Competitors and Evaluation Setup}

\begin{itemize}
  \item TUCH \cite{TUCH}: A model that is able to turn cross-view hashing into single-view hashing without multi-view embedding such that the information loss is minimized.
  \item CMDVH \cite{CMDVH}: A cross-modal deep variational hashing (CMDVH) method that learns non-linear transformations from image-text input pairs so that unified binary codes can be obtained. They also learn model-specific hashing networks in generative form, which is suitable for out-of-sample extension.
  \item DVSH \cite{DVSH}: The method performed end-to-end supervised metric-based training in the form of cosine hinge loss to obtain the binary codes, and learned modality-specific deep networks to obtain the hash functions.
  \item CorrAE \cite{Corr-AE}: The model is constructed by correlating hidden representations of two uni-modal auto-encoders, which is optimized by the correlation learning error between hidden representations of two modalities.
  \item CMNN \cite{Multimodal-hashing}: The method is to learn a similarity preserving network for cross-modalities through a coupled Siamese network with hinge loss.
  \item CAH \cite{CAH}: A model that is designed with a stacked auto-encoder architecture to jointly maximize the feature and semantic correlation across modalities.
  \item DCMH \cite{DCMH}: It is an end-to-end deep learning framework with a negative log likelihood criterion to preserve the similarity between real-value representations having the same class.
  \item HashGAN \cite{HashGAN}: An adversarial hashing network with attention mechanism to enhance the measurement of content similarities.
\end{itemize}

\textbf{Evaluation metrics}: For each data set, we perform two cross-modal retrieval tasks: image-to-text retrieval ($I \rightarrow T$) and text-to-image retrieval ($T \rightarrow I$), which search texts by a query image and search images by a query text, respectively. We use the mean average precision (mAP) to measure the performance of different retrieval methods. mAP is defined as the mean of all queries' average precision (AP), defined as $AP=\frac{1}{M} \sum_{r=1}^R prec(r) \odot rel(r)$ where $M$ is the number of relevant instances in the retrieved set, $prec(r)$ denotes the precision of the top $r$ retrieved set, and $rel(r)$ is an indicator of relevance of a given rank (which is set to 1 if relevant and 0 otherwise).

\subsection{Ablation Studies}

In this experiment, we provide analysis on the proposed loss function by comparing against ablations of the full objective, including the adversarial loss $\mathcal{L}_{GAN}$ alone and the cycle consistency loss $\mathcal{L}_{cyc}$ alone. In this experiment, following the cycle-GAN \cite{CycleGAN}, we adopt the FCN-score from \cite{Pix2Pix} as the automatic quantitative measure to evaluate the $text\rightarrow$ image task on Microsoft-COCO data set. The metric of fully-convolutional network (FCN) \cite{Fully-CNN} evaluates the how interpretable the generated images are in accordance to an off-the-shelf semantic segmentation algorithm. The FCN predicts a label map over a generated image, which can be compared against the input ground truth labels using the semantic segmentation metrics: per-pixel accuracy and per-class accuracy. The ablation studies on varied of our loss function are given in Table \ref{tab:FCN-score}. It can observed that muting either the GAN loss or the cycle-consistency loss can substantially degrade the accuracy results. We therefore conclude that both terms are critical to the regeneration results. This discovery is very consistent to that in cycle-GAN \cite{CycleGAN} which also claims that both cycle-consistent loss and GAN loss are indispensable in image translation.

\begin{table}[t]
  \centering
  \begin{tabular}{ |c|c|c|}
  \hline
  Loss & Per-class accuracy & Per-pixel accuracy \\
  \hline
  Cycle alone & 0.270 & 0.724  \\
  GAN alone & 0.611 & 0.126 \\
  CYC-DGH & \color{blue}$\mathbf{0.584}$ & \color{blue}$\mathbf{0.192}$ \\
  \hline
\end{tabular}
  \caption{FCN-scores for different variants of CYC-DGH, evaluated on Microsoft-COCO with 64 bits to regenerate the images.}\label{tab:FCN-score}
\end{table}

\subsection{Experimental Results}

In this section, we evaluate the model and algorithm from several aspects to demonstrate the superiority of the proposed CYC-DGH.

\subsubsection{Reconstruction Loss}

\begin{figure*}[t]
\centering
\begin{tabular}{ccc}
\includegraphics[height=3.5cm,width=5cm]{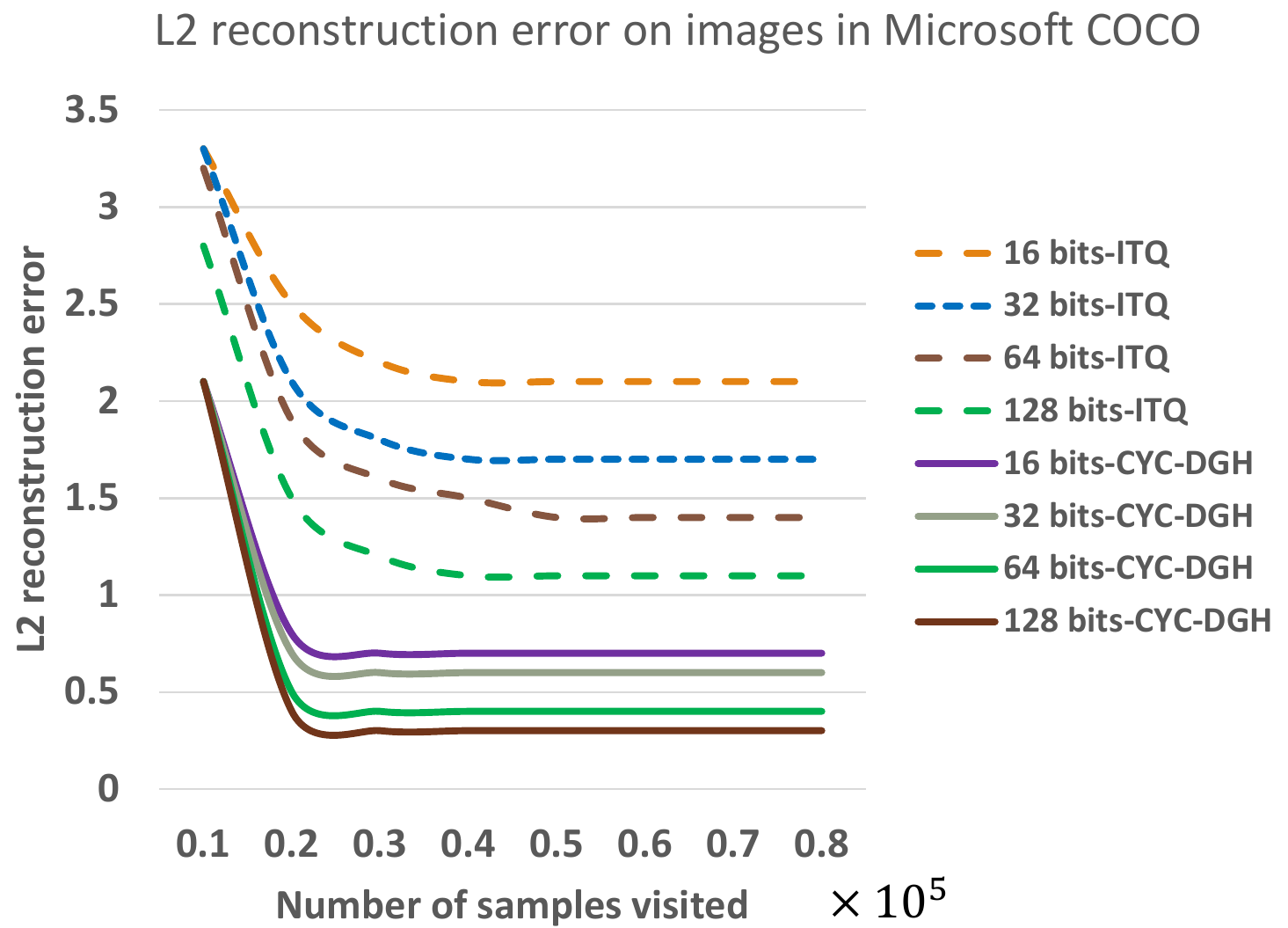}&
\includegraphics[height=3.5cm,width=5cm]{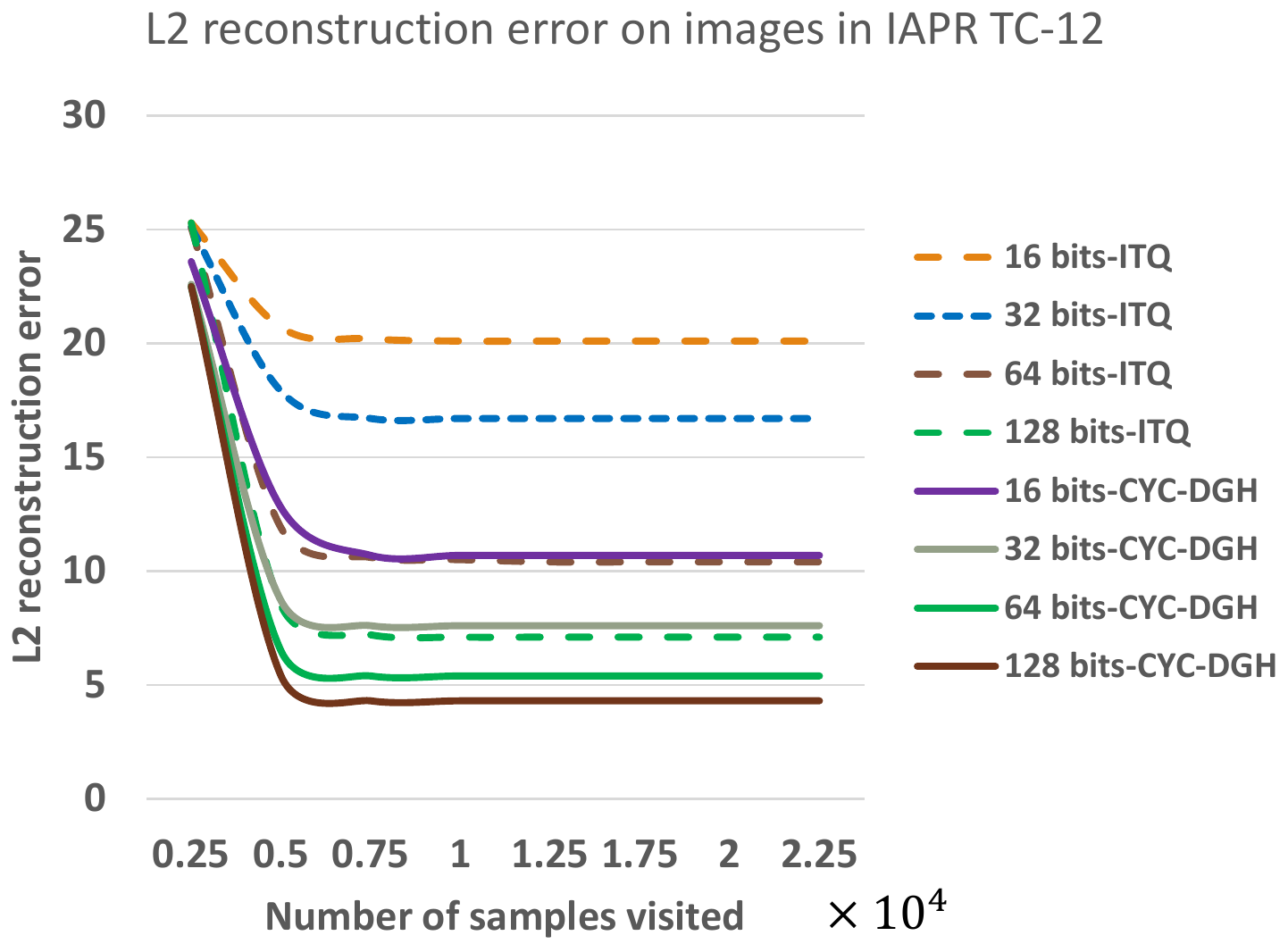}&
\includegraphics[height=3.5cm,width=5cm]{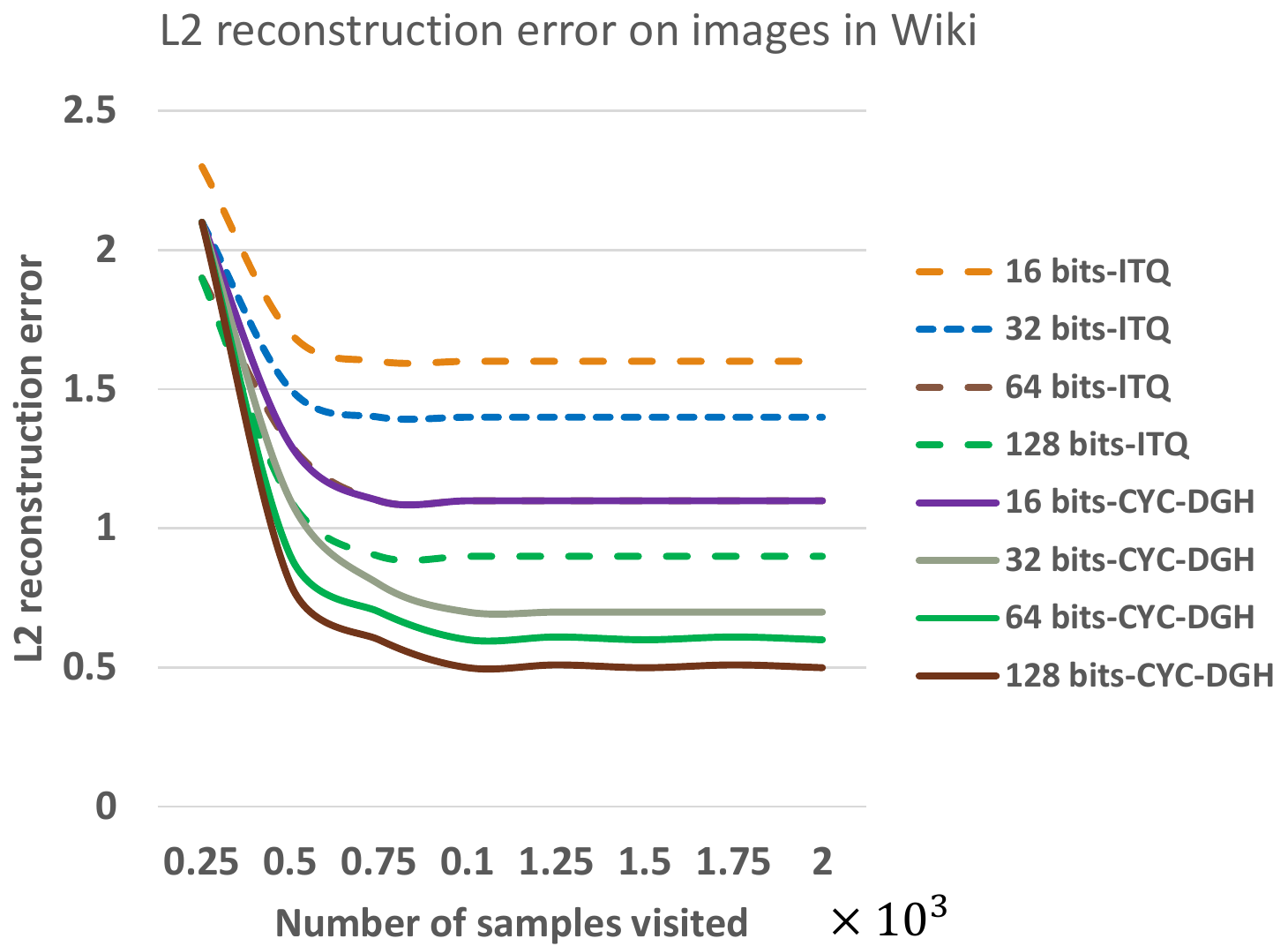}\\
\end{tabular}
\caption{The $L_2$ reconstruction error on images in three data sets.}
\label{fig:reconstruction-error}
\end{figure*}

To demonstrate the flexibility of generative modeling in reconstructing the inputs from the binary codes, we compare the $L_2$ reconstruction error to that of the ITQ \cite{ITQ}, showing the benefits of regenerating the inputs under the cycle-consistent constraint. Recall that our method has a generative model $\boldsymbol p(\boldsymbol x|\boldsymbol h)$, we can compute the regenerated input via $\hat{\boldsymbol x}=\arg\max \boldsymbol p(\boldsymbol x|\boldsymbol h)$, and then calculate the $L_2$ reconstruction loss of the regenerated input and the original $\boldsymbol x$ via $||\boldsymbol x- \hat{\boldsymbol x}||_2^2$. ITQ \cite{ITQ} trains by minimizing the binary quantization loss, that is, $\min ||B-XWR||_F^2$ (where $B=sign(XW)$, $X$ is the data matrix and $W$ is the encoding matrix), which is essentially $L_2$ reconstruction loss when the magnitude of the feature vectors is compatible with the radius of the binary code. Then, the ITQ \cite{ITQ} uses the covariance matrix $W$ formed by the eigenvalues and the orthogonal matrix $R$ to reconstruction the inputs, that is, $\min_{R, b}||\boldsymbol x_i - W R b_i||^2$. To have fair evaluation on the ITQ reconstruction capability for cross-modal retrieval, we use the covariance and orthogonal matrix learnt from the text domain to reconstruct the images: $||\boldsymbol x_u - W R b_v||_2$.  We plot the $L_2$ reconstruction error of our method and ITQ \cite{ITQ} on three data sets in Fig.\ref{fig:reconstruction-error}, where it shows the average $L_2$ reconstruction loss computed against the number of examples seen by the training process. The training time comparison with ITQ \cite{ITQ} are given in Table \ref{tab:train-time-COCO}, Table \ref{tab:train-time-IAPR}, and Table \ref{tab:train-time-Wiki}, respectively. Unlike ITQ \cite{ITQ} that iterative optimizes the quantization error under the orthogonal constraint, our proposed method is able to reconstruct the inputs through the Gaussian reconstruction without the orthogonality constraints, which can make the training more efficient. The lower reconstruction loss demonstrates the flexibility of the proposed CYC-DGH afforded by the re-parameterizations via stochastic neurons with cycle-consistent constraint. This is more apt to cross-modal retrieval and augment the correlation of image/text correspondence.

\begin{figure}[hbt]
\begin{tabular}{cc}
\includegraphics[height=3.5cm,width=4cm]{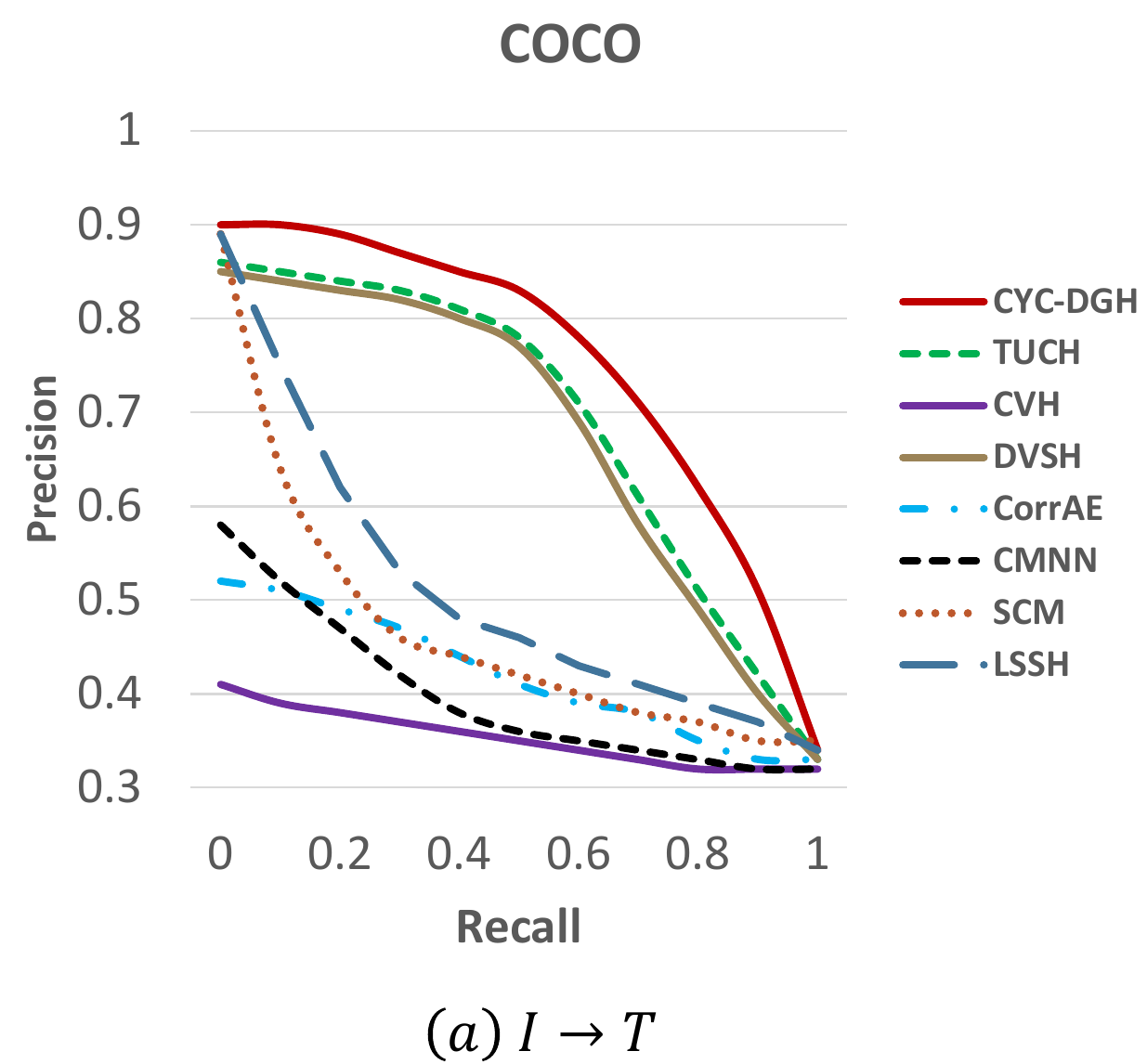}&
\includegraphics[height=3.5cm,width=4cm]{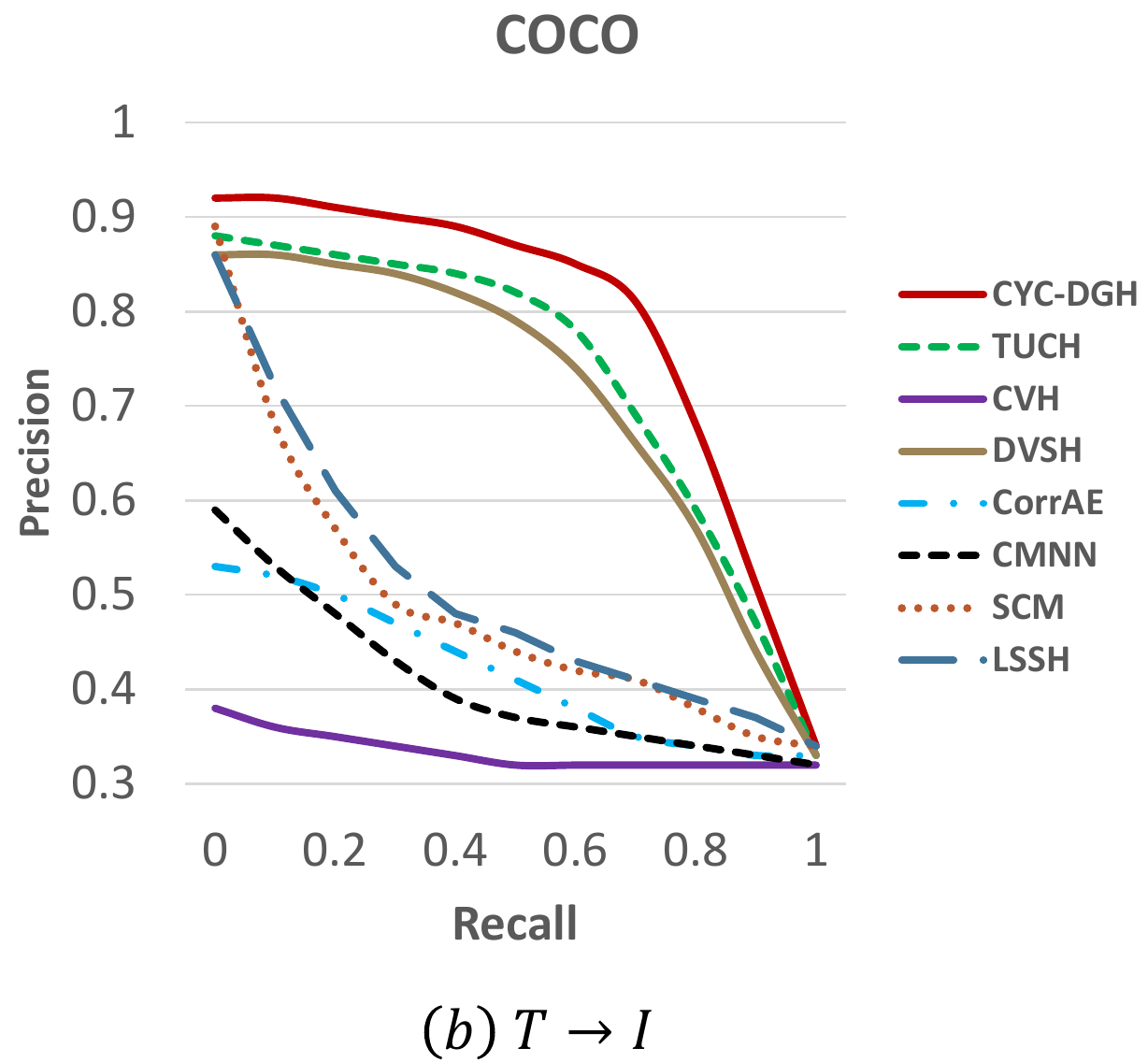}\\
\includegraphics[height=3.5cm,width=4cm]{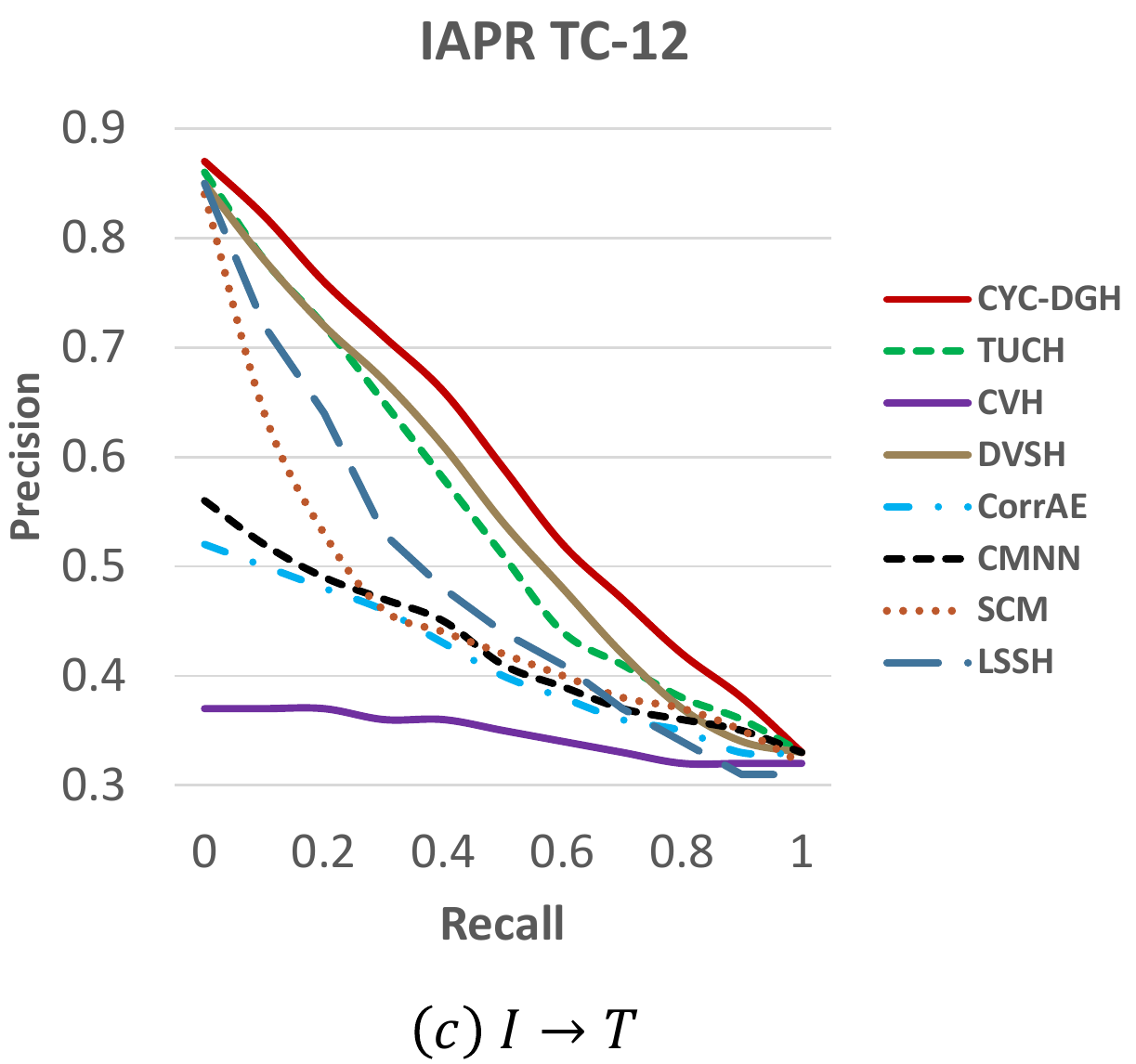}&
\includegraphics[height=3.5cm,width=4cm]{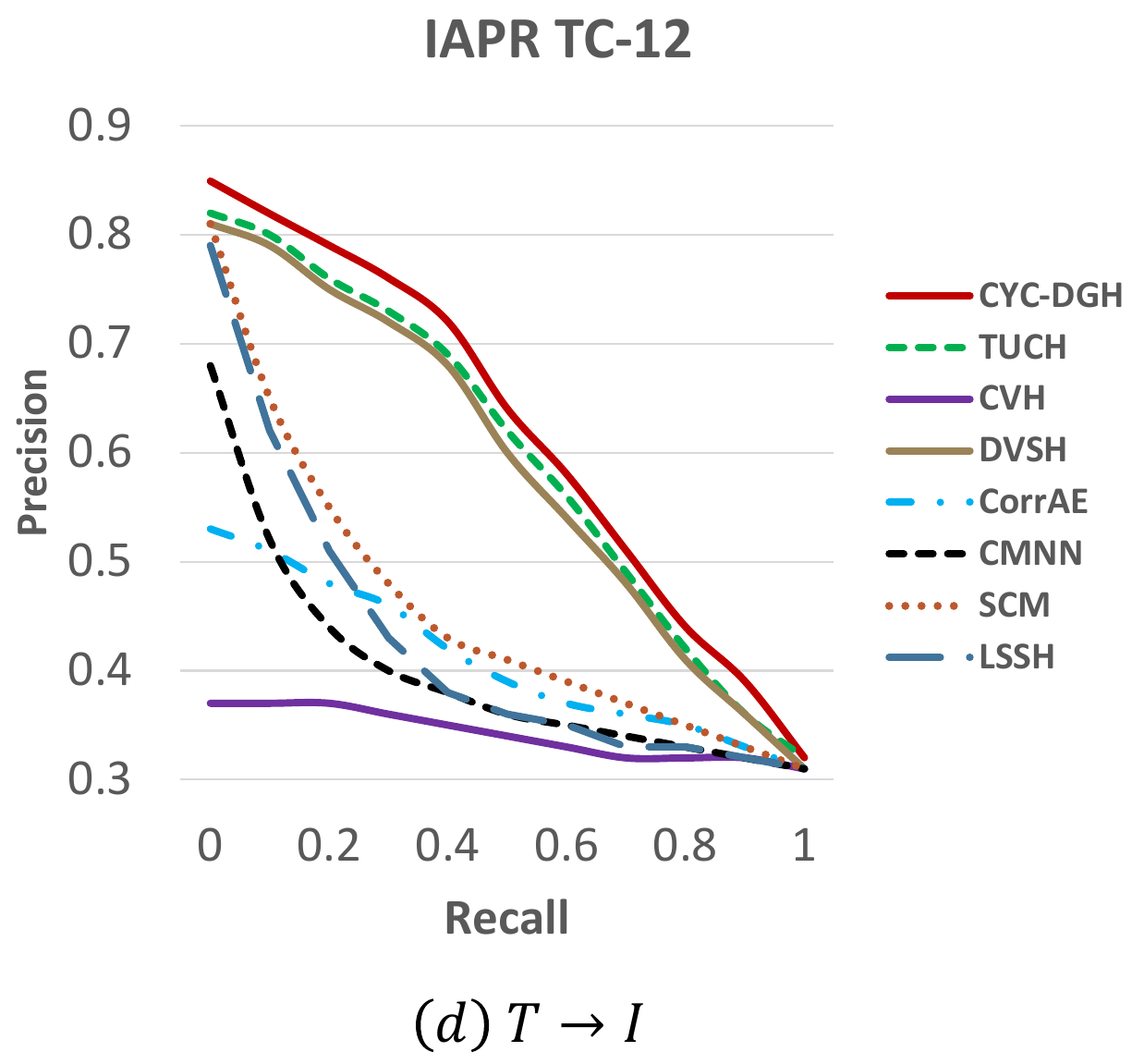}\\
\includegraphics[height=3.5cm,width=4cm]{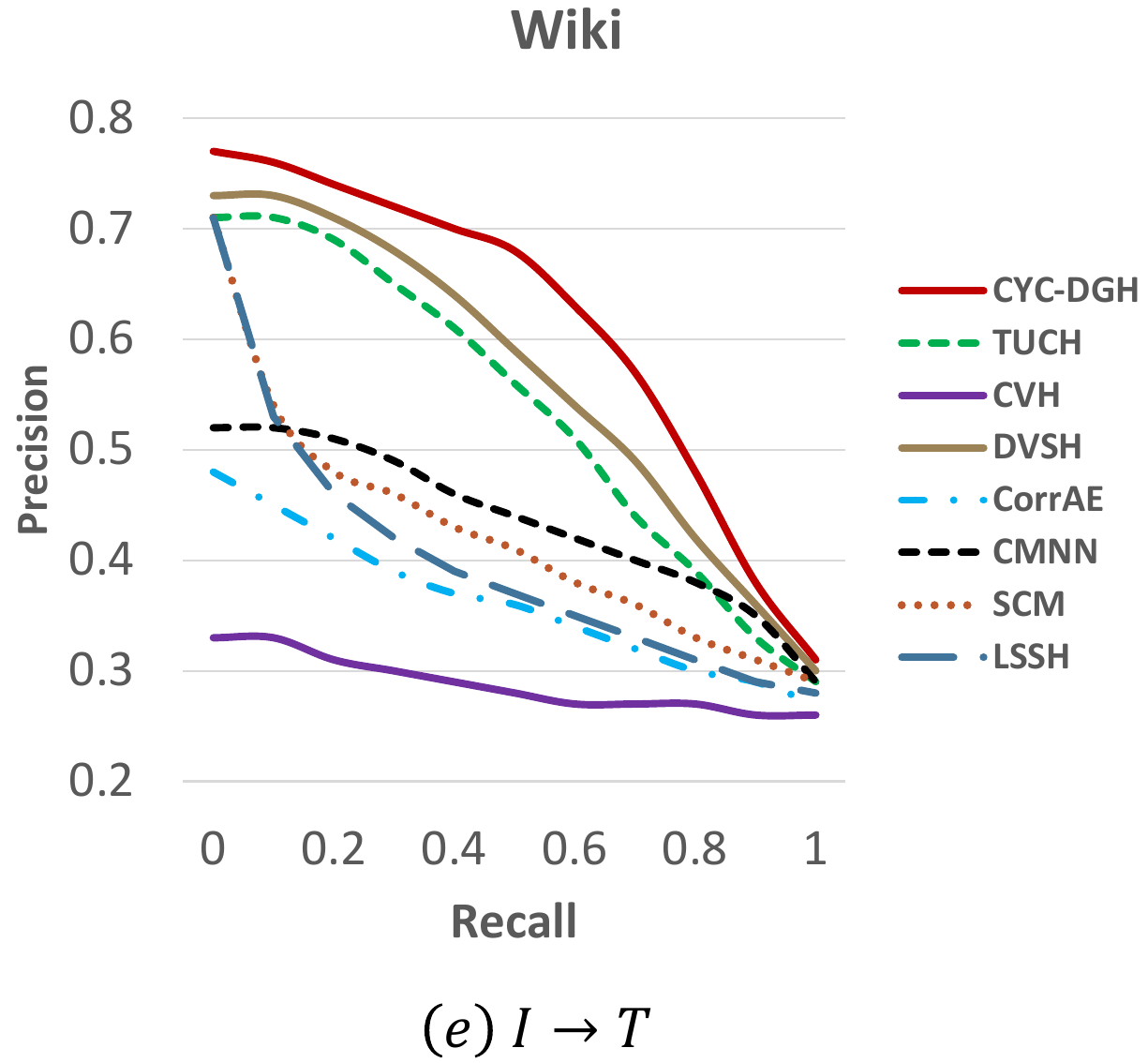}&
\includegraphics[height=3.5cm,width=4cm]{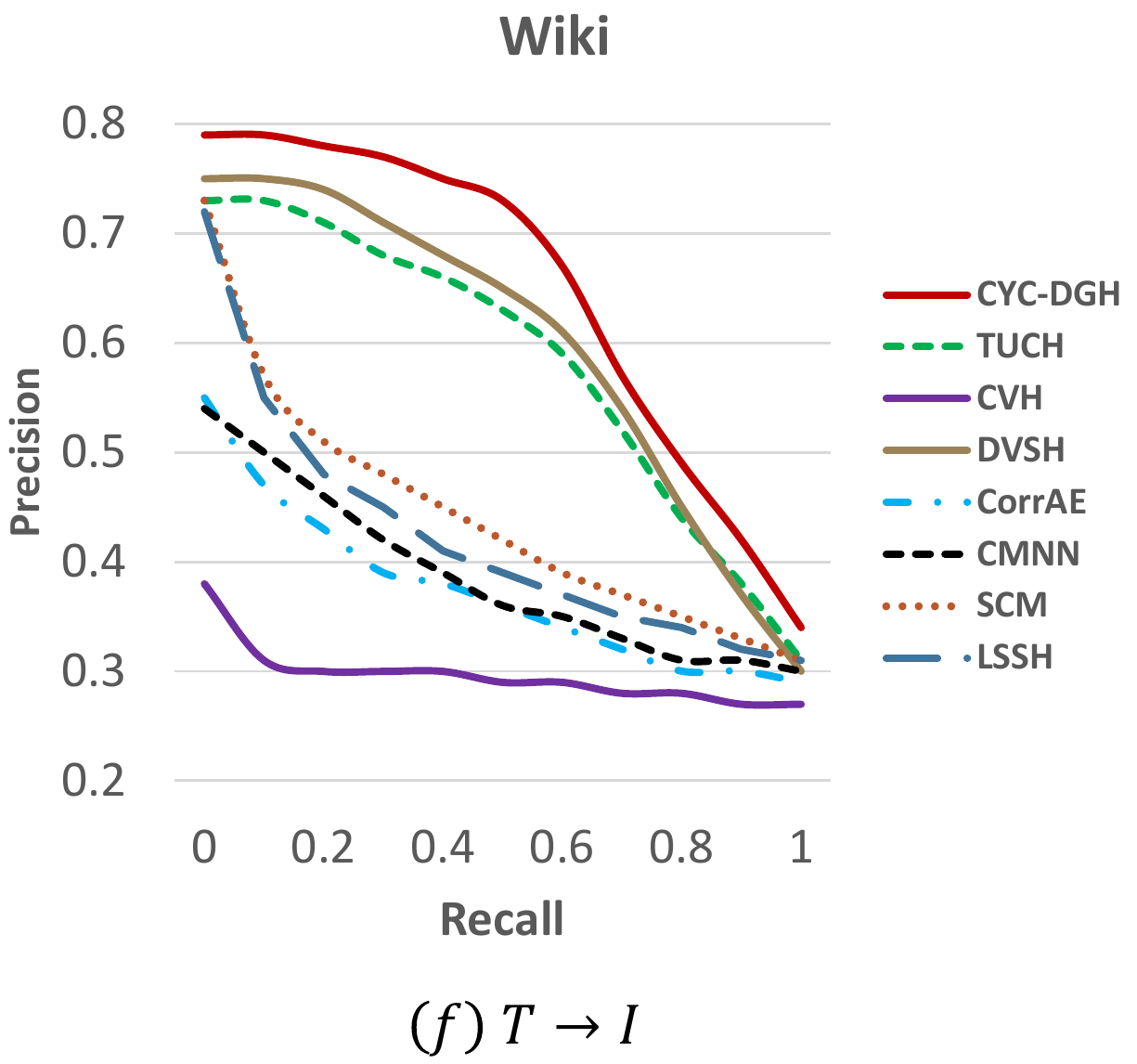}\\
\end{tabular}
\caption{The precision-recall curves of cross-modal retrieval on Microsoft COCO, IAPR TC-12, and Wiki @32 bits.}
\label{fig:precision-recall}
\end{figure}

\begin{table}[t]
  \centering
  \begin{tabular}{ |c|c|c|c|c|}
  \hline
  \multicolumn{5}{|c|}{Training time on Microsoft-COCO in seconds} \\
  \hline
  Method & 16 bits & 32 bits & 64 bits & 128 bits\\
  \hline
  CYC-DGH & 4.23 & 6.38 & 9.71 & 12.35 \\
  ITQ \cite{ITQ} & 22.74 & 38.36 & 51.91 & 67.23\\
  \hline
\end{tabular}
  \caption{Training time comparison on Microsoft-COCO.}\label{tab:train-time-COCO}
\end{table}

\begin{table}[t]
  \centering
  \begin{tabular}{ |c|c|c|c|c|}
  \hline
  \multicolumn{5}{|c|}{Training time on IAPR TC-12 in seconds} \\
  \hline
  Method & 16 bits & 32 bits & 64 bits & 128 bits\\
  \hline
  CYC-DGH & 3.92 & 5.84 & 9.11 & 11.05 \\
  ITQ \cite{ITQ} & 17.49 & 30.17 & 46.77 & 60.22\\
  \hline
\end{tabular}
  \caption{Training time comparison on IAPR TC-12.}\label{tab:train-time-IAPR}
\end{table}

\begin{table}[t]
  \centering
  \begin{tabular}{ |c|c|c|c|c|}
  \hline
  \multicolumn{5}{|c|}{Training time on Wiki in seconds} \\
  \hline
  Method & 16 bits & 32 bits & 64 bits & 128 bits\\
  \hline
  CYC-DGH & 2.03 & 3.18 & 5.32 & 7.65 \\
  ITQ \cite{ITQ} & 12.54 & 18.36 & 21.91 & 27.23\\
  \hline
\end{tabular}
  \caption{Training time comparison on Wiki.}\label{tab:train-time-Wiki}
\end{table}

\subsubsection{Comparison to State-of-the-art Cross-Modal Hashing Methods}

We compare our approach with state-of-the-art methods in terms of MAP, precision-recall curves and precision@top-$R$ returned curves in two cross-modal retrieval tasks: image query against textual database ($I \rightarrow T$), and textual query against image database ($T \rightarrow I$). The methods include unsupervised (CVH \cite{CVH}, LSSH \cite{LSSH}), and supervised (SCM \cite{SCM}, SePH \cite{Semantics-hash}). To have fair comparison with these non-deep-learning methods, we use the CNN features extracted at the FC7 layer for the images from the pre-trained model of CNN-F from \cite{M-Net}. Table \ref{tab:MAP-state-of-the-art} shows the mAP performance by the hamming ranking. For unsupervised methods such as CVH \cite{CVH} and LSSH \cite{LSSH}, their performance are less competitive to supervised. This is mainly because the cross-modal correlation can be achieved without the aid of semantics, and thus making the hashing learning of functions not discriminative. During retrieval, the supervised method of SePH \cite{Semantics-hash} can employ unified binary code learning across both the query set and the gallery set, and thus it achieves improved mAP values. However, SePH \cite{Semantics-hash} still requires the training sample in the form of aligned pairs, which would limit its application in practice. Also, it is clear to observe that our approach provides the best performance compared to these shallow cross-modal hashing methods.

\begin{table*}[hbt]
  \centering
  \caption{Mean Average Precision (MAP) comparison of state-of-the-art cross-modal hashing methods on three data sets.}\label{tab:MAP-state-of-the-art}
  {
  \begin{tabular}{l|l|c|c|c|c|c|c|c|c|c|c|c|c}
  \hline
  & & \multicolumn{4}{c|}{Microsoft COCO} & \multicolumn{4}{c|}{IAPR TC-12} & \multicolumn{4}{c}{Wiki}\\
  \cline{3-14}
   Task & Method & 16 bits & 32 bits & 64 bits & 128 bits & 16 bits & 32 bits & 64 bits & 128 bits & 16 bits & 32 bits & 64 bits & 128 bits \\
  \hline
   \multirow{5}{*}{$I \rightarrow T$}& CVH \cite{CVH} & 0.373 & 0.368 & 0.366 & 0.357 & 0.537 & 0.541 & 0.524 & 0.496 & 0.238 & 0.204 & 0.179 & 0.158 \\
   & SCM \cite{SCM} & 0.570 & 0.600 & 0.631 & 0.649 & 0.567 & 0.505 & 0.454 & 0.418 & 0.139 & 0.137 & 0.141 & 0.136 \\
   & LSSH \cite{LSSH} & - & - & - & - & 0.544 & 0.577 & 0.596 & 0.599 & 0.364 & 0.371 & 0.378 & 0.358\\
   & SePH \cite{Semantics-hash} & 0.581 & 0.613 & 0.625 & 0.634 & 0.618 & 0.645 & 0.650 & 0.678 & 0.414 & 0.435 & 0.437 & 0.447\\
   & CYC-DGH & \color{red}$\mathbf{0.722}$ & \color{red}$\mathbf{0.754}$ & \color{red}$\mathbf{0.781}$ & \color{red}$\mathbf{0.780}$ & \color{red}$\mathbf{0.771}$ & \color{red}$\mathbf{0.815}$ & \color{red}$\mathbf{0.832}$ & \color{red}$\mathbf{0.831}$ & \color{red}$\mathbf{0.794}$ & \color{red}$\mathbf{0.811}$ & \color{red}$\mathbf{0.813}$ & \color{red}$\mathbf{0.820}$ \\
  \hline
   \multirow{5}{*}{$T \rightarrow I$} & CVH \cite{CVH} & 0.373 & 0.369 & 0.365 & 0.371 & 0.568 & 0.578 & 0.561 & 0.536 & 0.388 & 0.336 & 0.257 & 0.230\\
   & SCM \cite{SCM} & 0.558 & 0.619 & 0.658 & 0.686 & 0.652 & 0.570 & 0.478 & 0.421 & 0.132 & 0.143 & 0.156 & 0.149\\
   & LSSH \cite{LSSH} & - & - & - & - & 0.487 & 0.526 & 0.555 & 0.572 & 0.606 & 0.626 & 0.638 & 0.638 \\
   & SePH \cite{Semantics-hash} & 0.613 & 0.650 & 0.672 & 0.693 & 0.610 & 0.634 & 0.640 & 0.673 & 0.701 & 0.699 & 0.710 & 0.715\\
   & CYC-DGH & \color{red}$\mathbf{0.761}$ & \color{red}$\mathbf{0.796}$ & \color{red}$\mathbf{0.834}$ & \color{red}$\mathbf{0.859}$ & \color{red}$\mathbf{0.772}$ & \color{red}$\mathbf{0.798}$ & \color{red}$\mathbf{0.837}$ & \color{red}$\mathbf{0.842}$ & \color{red}$\mathbf{0.811}$ & \color{red}$\mathbf{0.823}$ & \color{red}$\mathbf{0.826}$ & \color{red}$\mathbf{0.822}$ \\
  \hline
  \end{tabular}
  }
\end{table*}

\subsubsection{Comparison to Deep Cross-Modal Hashing Methods}

In this experiment, we compare our approach with recent deep cross-modal hashing competitors and show results in Table \ref{tab:MAP-values}. It can be seen that our method achieves the best results in terms of MAP values on different hashing bits over three data sets. This may be due to several reasons. First, the method of CAH \cite{CAH} still uses handcrafted image features as the input to their deep neural networks whereas our model starts learning from raw images. In the methods of DVSH \cite{DVSH} and CMDVH \cite{CMDVH}, the modality-specific hash functions are learned such that the non-linear relationship of samples from different modalities are exploited. Furthermore, CMDVH \cite{CMDVH} performed a shared binary code learning strategy to reduce the modality gap between the hash functions. However, explicitly learning modality-specific hash functions cannot render the hashing effective in cross-modal context. In other words, the learned hash functions are still limited in their specific modalities even in the joint learning with unified binary codes. In contrast, the proposed CYC-DGH eliminates the requirement of coupled training pairs in semantics, and the discriminative binary codes are produced through the enforced cycle-consistency loss. This cycle-consistent loss can uniquely correlate each pair with the same semantics and the modality heterogeneity can be addressed by the adversarial loss. The method of TUCH \cite{TUCH} utilizes the generative nets to convert one modality data into the target modality so as to minimize the information loss caused by the respective hashing embedding. However, converting data into a different modality is unable to maintain the relationship in the original source data and thus cannot achieve very comparable results to CMDVH \cite{CMDVH} and our method. In contrast, the proposed CYC-DGH effectively address the information loss by proposing to regenerating the inputs through the binary codes, and the hash function learning as well as regeneration process are jointly achieved in the full loss function.

The precision-recall curves with 32 bits on for the two cross-modal tasks on three benchmarks are shown in Fig.\ref{fig:precision-recall}, respectively. It can be seen that CYC-DGH achieves the best performance at two asks on all recall levels. This is mainly because the removal of paired training correspondence can still be supplemented by the effective cycle-consistency loss, and the adversarial training is able to reduce the modality heterogeneity gap. Moreover, another major challenge in cross-modal retrieval is the information loss caused by the binary embedding which is combated by our approach with input regeneration from binary codes.

\begin{figure}[t]
\begin{tabular}{cc}
\includegraphics[height=3.5cm,width=4cm]{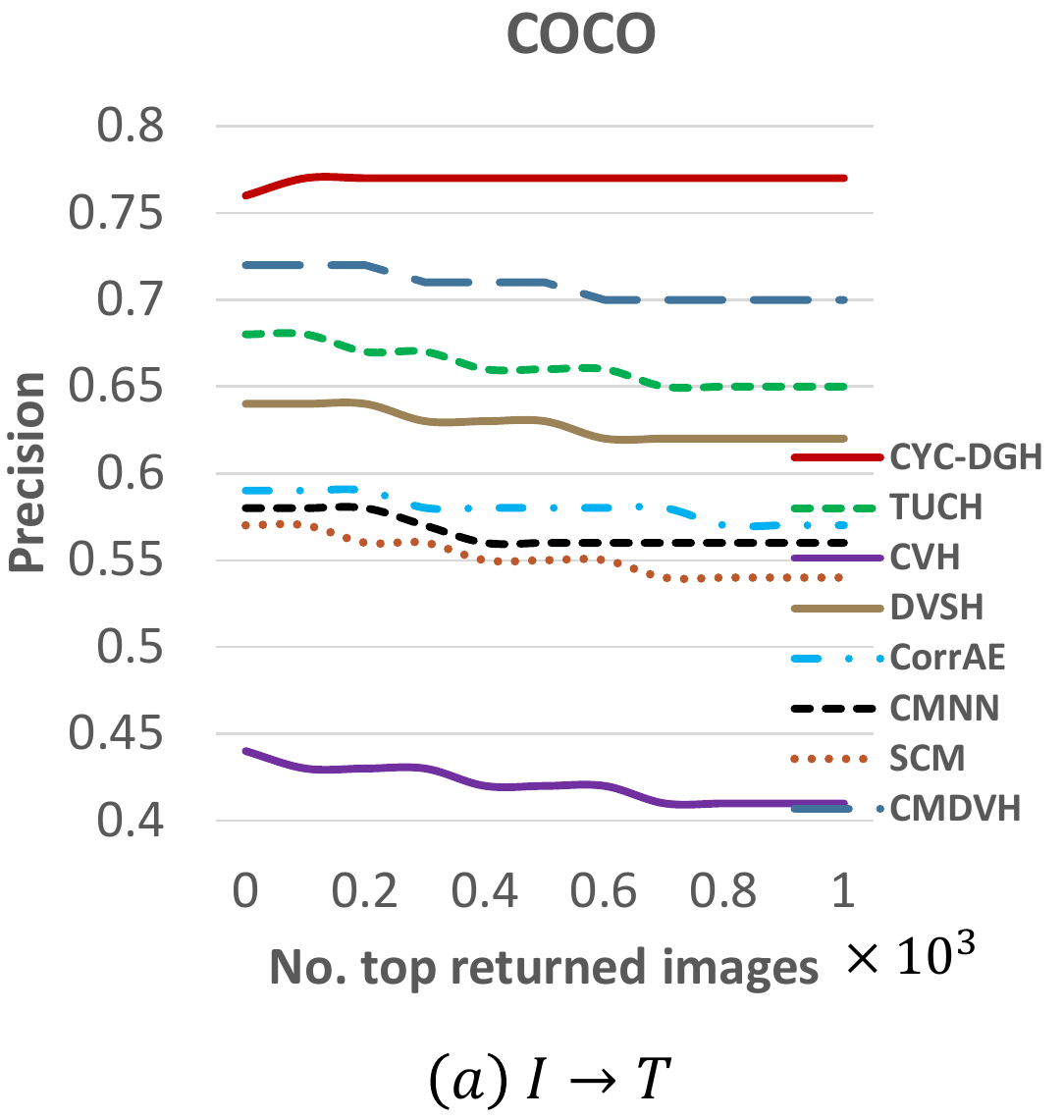}&
\includegraphics[height=3.5cm,width=4cm]{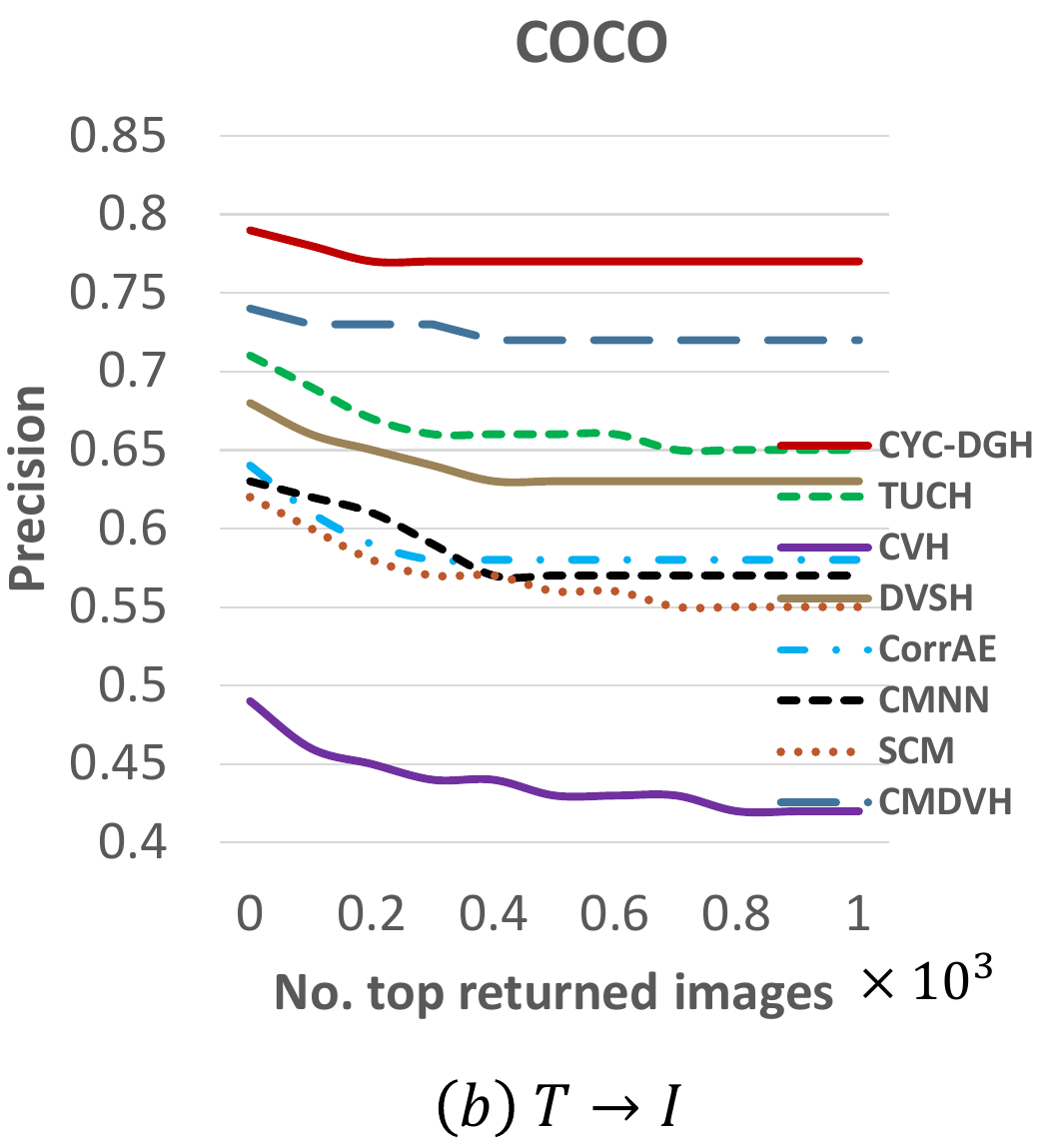}\\
\includegraphics[height=3.5cm,width=4cm]{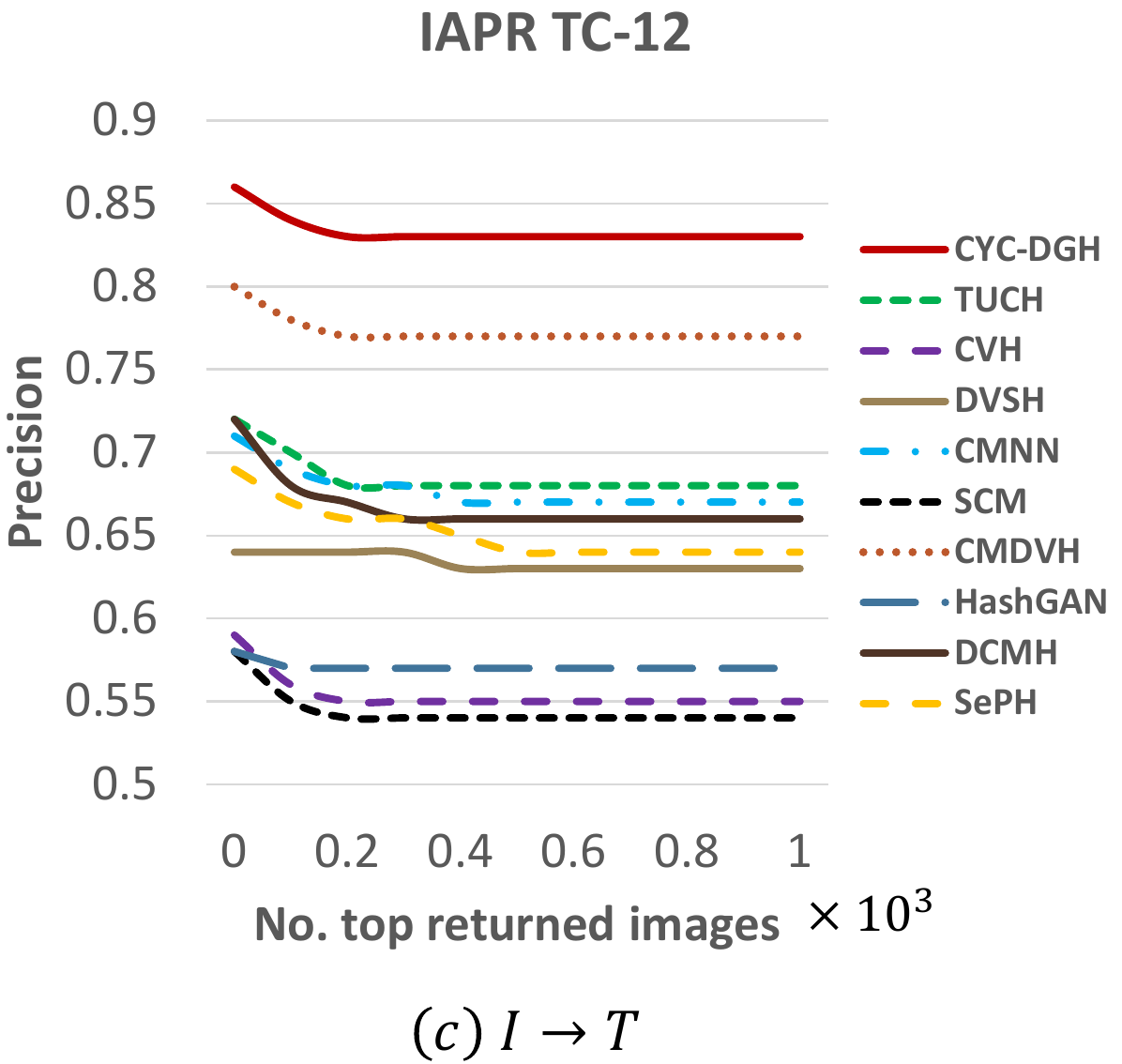}&
\includegraphics[height=3.5cm,width=4cm]{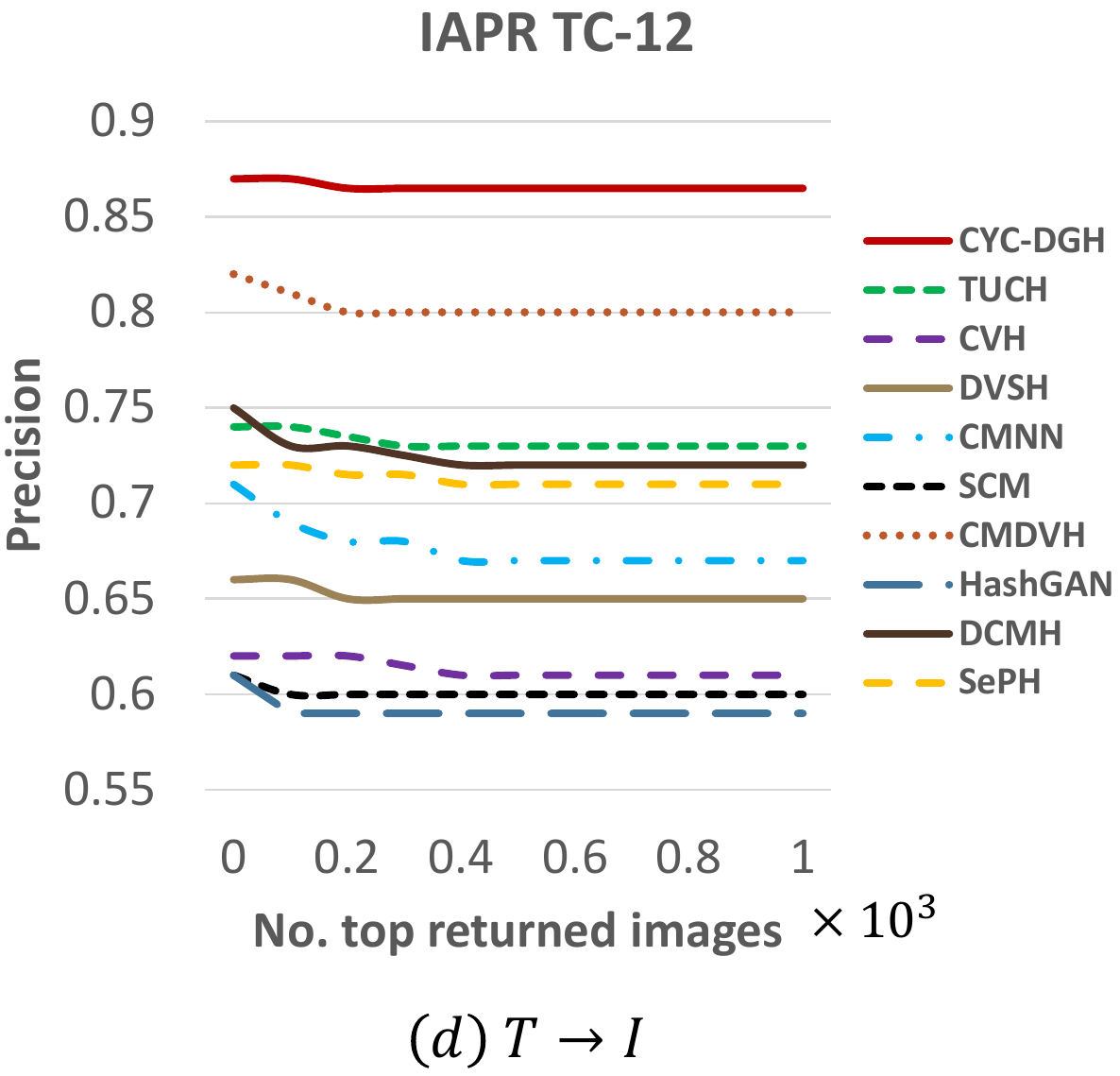}\\
\includegraphics[height=3.5cm,width=4cm]{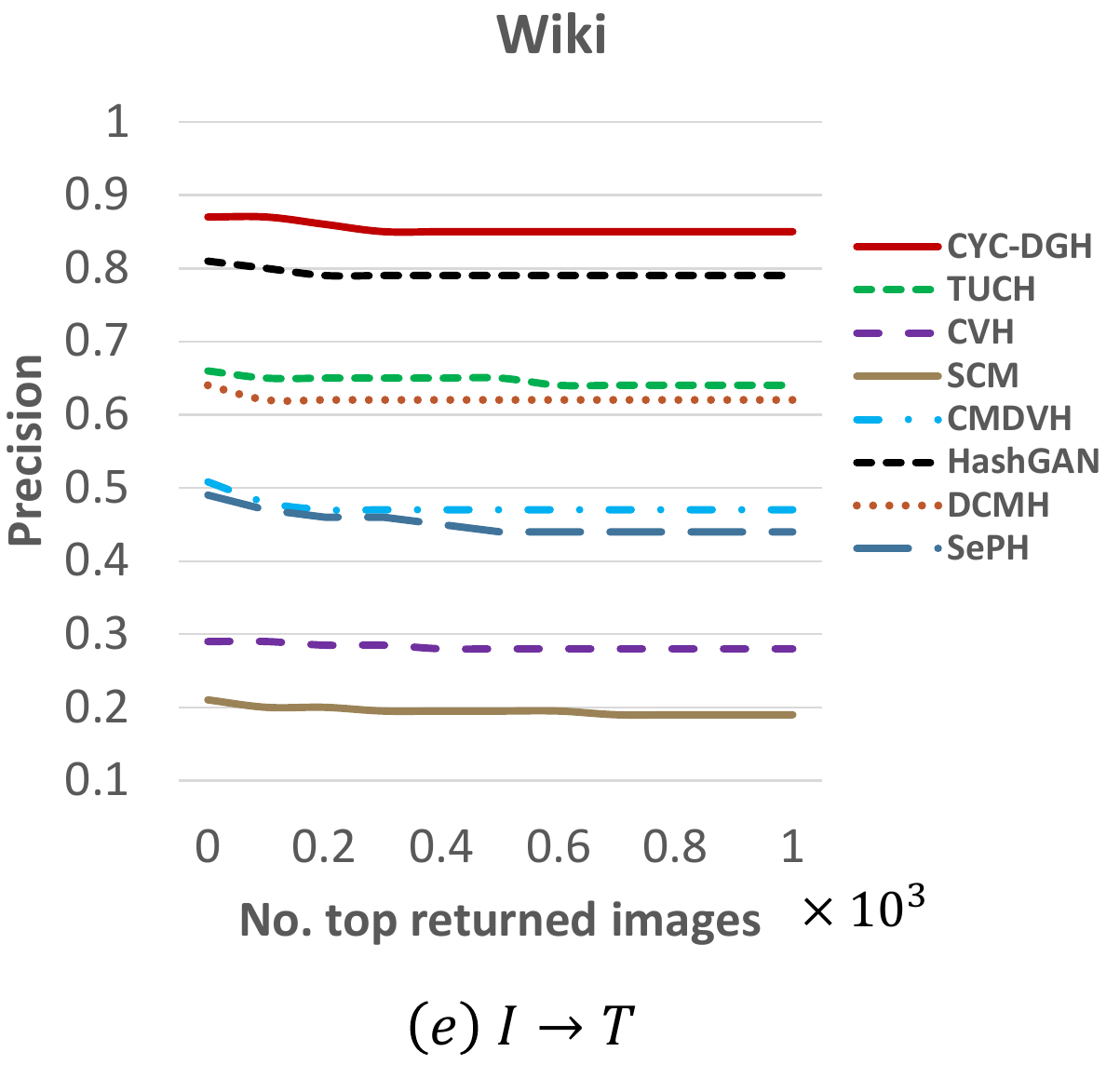}&
\includegraphics[height=3.5cm,width=4cm]{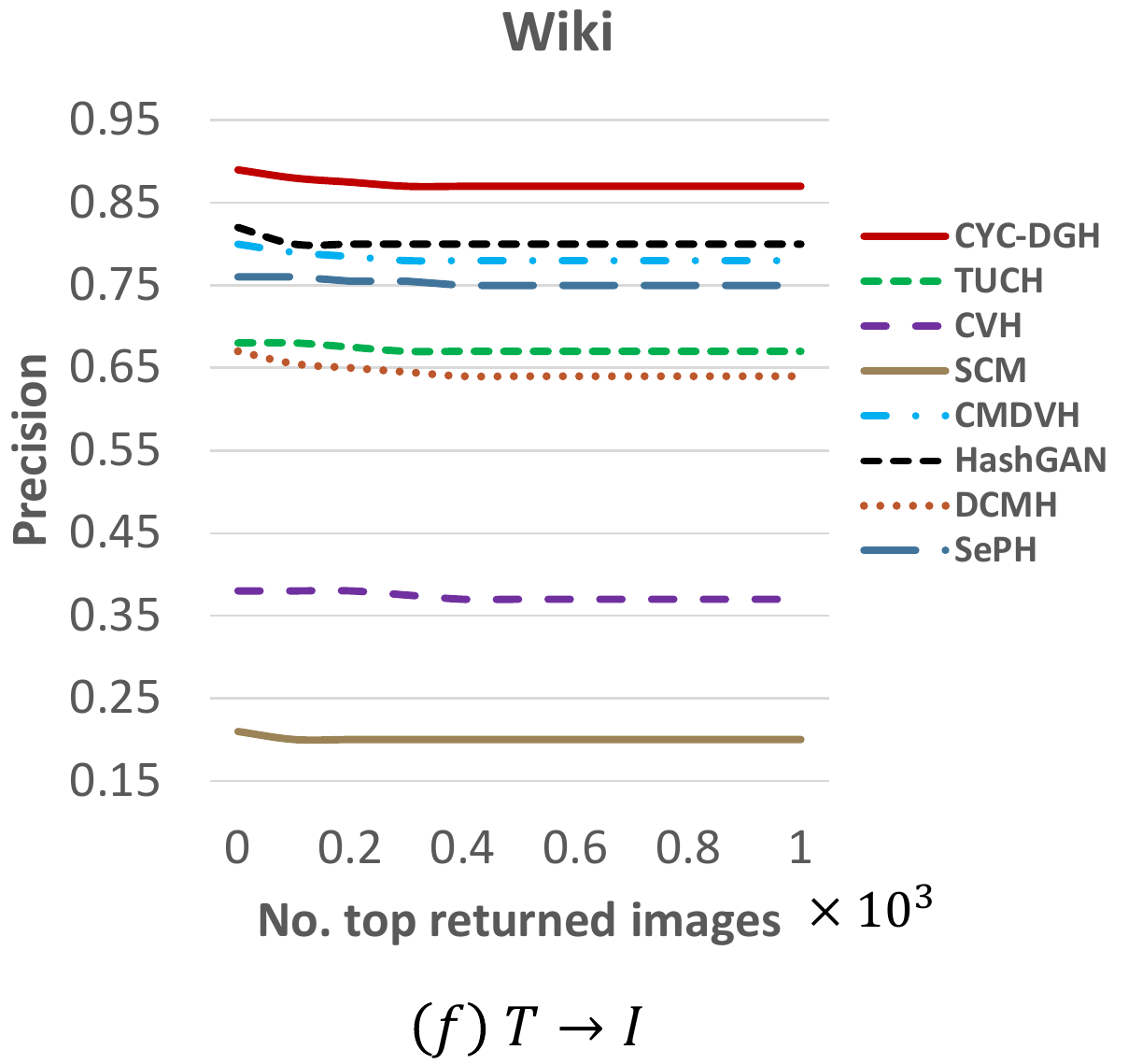}\\
\end{tabular}
\caption{The precision@top-R return curves of cross-modal retrieval on Microsoft COCO, IAPR TC-12, and Wiki @32 bits.}
\label{fig:precision-top-return}
\end{figure}

Figure. \ref{fig:precision-top-return} shows the precision@top-$R$ return curves of all comparison methods with 32 bits on three data sets. It displays how the precision changes against the number of $R$ of top-retrieved results. CYC-DGH outperforms all the competitors and shows consistent effectiveness against the increased number of top-retrieved items.

\begin{table*}[hbt]
  \centering
  \caption{Mean Average Precision (MAP) comparison of deep cross-modal hashing methods on three data sets.}\label{tab:MAP-values}
  {
  \begin{tabular}{l|l|c|c|c|c|c|c|c|c|c|c|c|c}
  \hline
  & & \multicolumn{4}{c|}{Microsoft COCO} & \multicolumn{4}{c|}{IAPR TC-12} & \multicolumn{4}{c}{Wiki}\\
  \cline{3-14}
   Task & Method & 16 bits & 32 bits & 64 bits & 128 bits & 16 bits & 32 bits & 64 bits & 128 bits & 16 bits & 32 bits & 64 bits & 128 bits \\
  \hline
   \multirow{9}{*}{$I \rightarrow T$} & TUCH \cite{TUCH} & 0.628 & 0.714 & 0.735 & 0.766 & 0.595 & 0.637 & 0.690 & 0.714 & 0.578 & 0.621 & 0.637 & 0.660 \\
   & CMDVH \cite{CMDVH} & 0.669 & 0.697 & 0.721 & 0.769 & 0.720 & 0.773 & 0.800 & 0.790 & 0.424 & 0.443 & 0.452 & 0.444\\
   & DVSH \cite{DVSH} & 0.587 & 0.713 & 0.739 & 0.755 & 0.570 & 0.632 & 0.696 & 0.723 & - & - & - & - \\
   & CMNN \cite{Multimodal-hashing} & 0.556 & 0.560 & 0.585 & 0.594 & 0.516 & 0.542 & 0.577 & 0.600 & - & - & - & - \\
   & CorrAE \cite{Corr-AE} & 0.550 & 0.556 & 0.569 & 0.581 & 0.495 & 0.525 & 0.557 & 0.589 & 0.507 & 0.548 & 0.569 & 0.610 \\
   & CAH \cite{CAH} & - & - & - & - & 0.559 & 0.590 & 0.603 & 0.609 & 0.492 & 0.502 & 0.542 & 0.560 \\
   & DCMH \cite{DCMH} & - & - & - & - & 0.624 & 0.635 & 0.676 & 0.671 & 0.569 & 0.588 & 0.631 & 0.633 \\
   & HashGAN \cite{HashGAN} & - & - & - & - & 0.529 & 0.528 & 0.544 & 0.546 & 0.756 & 0.772 & 0.772 & 0.774 \\
   & CYC-DGH & \color{red}$\mathbf{0.722}$ & \color{red}$\mathbf{0.754}$ & \color{red}$\mathbf{0.781}$ & \color{red}$\mathbf{0.780}$ & \color{red}$\mathbf{0.771}$ & \color{red}$\mathbf{0.815}$ & \color{red}$\mathbf{0.832}$ & \color{red}$\mathbf{0.831}$ & \color{red}$\mathbf{0.794}$ & \color{red}$\mathbf{0.811}$ & \color{red}$\mathbf{0.813}$ & \color{red}$\mathbf{0.820}$ \\
  \hline
   \multirow{9}{*}{$T \rightarrow I$} & TUCH \cite{TUCH} & 0.649 & 0.760 & 0.786 & 0.812 & 0.624 & 0.656 & 0.688 & 0.709 & 0.598 & 0.627 & 0.655 & 0.679\\
   & CMDVH \cite{CMDVH} & 0.691 & 0.735 & 0.768 & 0.765 & 0.735 & 0.774 & 0.804 & 0.811 & 0.727 & 0.733 & 0.738 & 0.737\\
   & DVSH \cite{DVSH} & 0.591 & 0.737 & 0.758 & 0.767 & 0.604 & 0.640 & 0.681 & 0.675 & - & - & - & - \\
   & CMNN \cite{Multimodal-hashing} & 0.579 & 0.598 & 0.620 & 0.645 & 0.512 & 0.540 & 0.549 & 0.565 & - & - & - & - \\
   & CorrAE \cite{Corr-AE} & 0.559 & 0.581 & 0.611 & 0.626 & 0.498 & 0.519 & 0.533 & 0.549 & 0.523 & 0.561 & 0.582 & 0.617 \\
   & CAH \cite{CAH} & - & - & - & - & 0.582 & 0.621 & 0.638 & 0.647 & 0.527 & 0.534 & 0.570 & 0.596 \\
   & DCMH \cite{DCMH} & - & - & - & - & 0.648 & 0.672 & 0.698 & 0.707 & 0.589 & 0.611 & 0.651 & 0.662 \\
   & HashGAN \cite{HashGAN} & - & - & - & - & 0.536 & 0.557 & 0.565 & 0.571 & 0.792 & 0.806 & 0.807 & 0.809 \\
   & CYC-DGH & \color{red}$\mathbf{0.761}$ & \color{red}$\mathbf{0.796}$ & \color{red}$\mathbf{0.834}$ & \color{red}$\mathbf{0.859}$ & \color{red}$\mathbf{0.772}$ & \color{red}$\mathbf{0.798}$ & \color{red}$\mathbf{0.837}$ & \color{red}$\mathbf{0.842}$ & \color{red}$\mathbf{0.811}$ & \color{red}$\mathbf{0.823}$ & \color{red}$\mathbf{0.826}$ & \color{red}$\mathbf{0.822}$ \\
  \hline
  \end{tabular}
  }
\end{table*}

\section{Conclusion and Future Work}\label{sec:con}

In this paper, we present a novel deep generative hashing framework towards cross-modal retrieval that is able to produce hashing learning functions without the need of paired training samples. The proposed model, namely Cycle-consistent Deep Generative Hashing (CYC-DGH), builds adversarial training across modality to reduce the heterogeneity, which is augmented with cycle-consistent constraint to uniquely maximize the correlation of the input-output without the semantic labeling. Besides, we introduce the generative model into the hashing learning, which jointly performs binary code learning as well as input generation from binary codes so as to minimize the information loss to great extent. Extensive empirical evidences including comparison with competitors and ablation studies show that our CYC-DGH approach advance the state-of-the-arts on image to text (and text to image) retrieval tasks, over three benchmarks. In the future, we plan to explore more powerful generative models to further improve the generation capability of our method.

\bibliographystyle{IEEEtran}\small
\bibliography{allbib}

\begin{IEEEbiography}{Lin Wu}
was awarded a PhD from The University of New South Wales, Sydney, Australia in 2014.  So far, she has published 40 academic papers, such as CVPR, ACM Multimedia, IJCAI, ACM SIGIR, IEEE-TIP, IEEE-TNNLS, IEEE-TCYB, Neural Networks, Pattern Recognition.  She also regularly served as the program committee member for numerous international conferences and invited journal reviewer for IEEE TIP, IEEE TNNLS, IEEE TCSVT, IEEE TMM and Pattern Recognition.
\end{IEEEbiography}
\begin{IEEEbiography}{Yang Wang}
obtained his PhD degree at The University of New South Wales, Kensington, Sydney, Australia in 2015.  He has published 40 research papers together with a book chapter, most of which have appeared at the competitive venues, including IEEE TIP, IEEE TNNLS, IEEE TCYB, IEEE TKDE, Pattern Recognition, Neural Networks, ACM Multimedia, ACM SIGIR, IJCAI, IEEE ICDM, ACM CIKM and VLDB Journal.  He regularly served as the invited journal reviewer for more than 15 leading journals such as IEEE TPAMI, IEEE TIP, IEEE TNNLS, IEEE TKDE and Machine Learning Springer, IEEE TMM etc.
\end{IEEEbiography}



\begin{IEEEbiography}{Ling Shao}
is a Director for Inception Institute of Artificial Intelligence (IIAI), Abu Dhabi, United Arab Emirates, and a Professor at University of East Anglia, United Kingdom. He has authored/co-authored over 250 papers in refereed journals/conferences such as IEEE TPAMI, TIP, TNNLS, IJCV, ICCV, CVPR, ECCV, IJCAI and ACM MM, and holds over 10 EU/US patents. Prof Shao is an Associate Editor of IEEE Transactions on Image Processing, IEEE Transactions on Neural Networks and Learning Systems, IEEE Transactions on Circuits and Systems for Video Technology.  Prof. Shao is the General Chair of British Machine Vision Conference (BMVC) 2018.
\end{IEEEbiography}

%

\ifCLASSOPTIONcaptionsoff
  \newpage
\fi

\end{document}